\documentclass[10pt,journal,compsoc]{IEEEtran}
\newif\ifpeerreview

\peerreviewfalse

\usepackage[nocompress]{cite}
\usepackage{url}
\usepackage{amsmath,amssymb,graphicx}
\usepackage{tabularx}
\usepackage{rotating}
\usepackage{graphicx}
\usepackage{booktabs}
\usepackage{siunitx}
\usepackage{multirow}
\usepackage{ntheorem}
\usepackage{cases}
\usepackage{array}
\usepackage{glossaries-extra}
\usepackage{xspace}
\usepackage{wrapfig}
\usepackage{biolinum}
\usepackage[table]{xcolor}
\usepackage{makecell}
\usepackage{lipsum} 
\usepackage[switch]{lineno}
\usepackage{color}
\usepackage{xcolor}
\usepackage{tikz}
\usepackage{sankey}
\usetikzlibrary{backgrounds,calc,positioning}

\usepackage[table]{xcolor}
\usepackage{colortbl,booktabs,array,siunitx}
\usepackage{pgf}

\newcommand{\na}{--}

\newcommand{\heatup}[3]{%
  \begingroup
  \pgfmathsetmacro{\p}{max(0,min(1,((#1)-(#2))/((#3)-(#2))))}%
  \pgfmathsetmacro{\R}{1.00 - 0.35*\p}%
  \pgfmathsetmacro{\G}{0.78 + 0.22*\p}%
  \pgfmathsetmacro{\B}{0.78 - 0.30*\p}%
  \edef\x{\endgroup
    \noexpand\cellcolor[rgb]{\R,\G,\B}\noexpand\color{black}\noexpand\num{#1}}%
  \x
}

\newcommand{\heatdown}[3]{%
  \begingroup
  \pgfmathsetmacro{\p}{max(0,min(1,((#3)-(#1))/((#3)-(#2))))}%
  \pgfmathsetmacro{\R}{1.00 - 0.35*\p}%
  \pgfmathsetmacro{\G}{0.78 + 0.22*\p}%
  \pgfmathsetmacro{\B}{0.78 - 0.30*\p}%
  \edef\x{\endgroup
    \noexpand\cellcolor[rgb]{\R,\G,\B}\noexpand\color{black}\noexpand\num{#1}}%
  \x
}

\newcommand{\psnrcell}[1]{\heatup{#1}{20}{40}}
\newcommand{\ssimcell}[1]{\heatup{#1}{0.40}{0.95}}
\newcommand{\lpipscell}[1]{\heatdown{#1}{0.05}{0.60}}
\newcommand{\niqecell}[1]{\heatdown{#1}{2.0}{10.0}}

\newcommand{\heatupbox}[3]{%
  \begingroup
  \pgfmathsetmacro{\p}{max(0,min(1,((#1)-(#2))/((#3)-(#2))))}%
  \pgfmathsetmacro{\R}{1.00 - 0.35*\p}%
  \pgfmathsetmacro{\G}{0.78 + 0.22*\p}%
  \pgfmathsetmacro{\B}{0.78 - 0.30*\p}%
  \edef\x{\endgroup
    \noexpand\colorbox[rgb]{\R,\G,\B}{\noexpand\color{black}\noexpand\raisebox{0pt}[1.45ex][0.35ex]{\noexpand\num[round-mode=places,round-precision=4]{#1}}}}%
  \x
}

\newcommand{\heatdownbox}[3]{%
  \begingroup
  \pgfmathsetmacro{\p}{max(0,min(1,((#3)-(#1))/((#3)-(#2))))}%
  \pgfmathsetmacro{\R}{1.00 - 0.35*\p}%
  \pgfmathsetmacro{\G}{0.78 + 0.22*\p}%
  \pgfmathsetmacro{\B}{0.78 - 0.30*\p}%
  \edef\x{\endgroup
    \noexpand\colorbox[rgb]{\R,\G,\B}{\noexpand\color{black}\noexpand\raisebox{0pt}[1.45ex][0.35ex]{\noexpand\num[round-mode=places,round-precision=4]{#1}}}}%
  \x
}

\newcommand{\maybeheatupbox}[3]{\ifstrequal{#1}{\na}{\na}{\heatupbox{#1}{#2}{#3}}}
\newcommand{\maybeheatdownbox}[3]{\ifstrequal{#1}{\na}{\na}{\heatdownbox{#1}{#2}{#3}}}

\newcommand{\stackheatup}[4]{%
  \begingroup
  \setlength{\fboxsep}{0.5pt}%
  \vtop{\offinterlineskip\halign{\hfil##\hfil\cr
    \maybeheatupbox{#1}{#3}{#4}\cr
    \maybeheatupbox{#2}{#3}{#4}\cr}}%
  \endgroup
}

\newcommand{\stackheatdown}[4]{%
  \begingroup
  \setlength{\fboxsep}{0.5pt}%
  \vtop{\offinterlineskip\halign{\hfil##\hfil\cr
    \maybeheatdownbox{#1}{#3}{#4}\cr
    \maybeheatdownbox{#2}{#3}{#4}\cr}}%
  \endgroup
}

\newcommand{\tristackheatup}[5]{%
  \begingroup
  \setlength{\fboxsep}{0.5pt}%
  \vtop{\offinterlineskip\halign{\hfil##\hfil\cr
    \maybeheatupbox{#1}{#4}{#5}\cr
    \maybeheatupbox{#2}{#4}{#5}\cr
    \maybeheatupbox{#3}{#4}{#5}\cr}}%
  \endgroup
}

\newcommand{\tristackheatdown}[5]{%
  \begingroup
  \setlength{\fboxsep}{0.5pt}%
  \vtop{\offinterlineskip\halign{\hfil##\hfil\cr
    \maybeheatdownbox{#1}{#4}{#5}\cr
    \maybeheatdownbox{#2}{#4}{#5}\cr
    \maybeheatdownbox{#3}{#4}{#5}\cr}}%
  \endgroup
}

\newcommand{\stackpsnrcell}[2]{\stackheatup{#1}{#2}{20}{40}}
\newcommand{\stackssimcell}[2]{\stackheatup{#1}{#2}{0.40}{0.95}}
\newcommand{\stacklpipscell}[2]{\stackheatdown{#1}{#2}{0.05}{0.60}}
\newcommand{\stackniqecell}[2]{\stackheatdown{#1}{#2}{2.0}{10.0}}
\newcommand{\tripsnrcell}[3]{\tristackheatup{#1}{#2}{#3}{20}{40}}
\newcommand{\trissimcell}[3]{\tristackheatup{#1}{#2}{#3}{0.40}{0.95}}
\newcommand{\trilpipscell}[3]{\tristackheatdown{#1}{#2}{#3}{0.05}{0.60}}
\newcommand{\triniqecell}[3]{\tristackheatdown{#1}{#2}{#3}{2.0}{10.0}}

\newcommand{\imgicon}[1]{\includegraphics[height=6pt]{#1}}

\newabbreviation{PSF}{PSF}{Point Spread Function}
\global\long\def\PSF{\gls{PSF}\xspace}
\newabbreviation{VLMs}{VLMs}{Vision-Language Models}
\global\long\def\VLMs{\gls{VLMs}\xspace}
\newabbreviation{VLM}{VLM}{Vision-Language Model}
\global\long\def\VLM{\gls{VLM}\xspace}
\newabbreviation{ImagingBench}{ImagingBench}{Imaging Benchmark}
\global\long\def\ImagingBench{\gls{ImagingBench}\xspace}
\newabbreviation{CGH}{CGH}{Computer Generated Holography}
\global\long\def\CGH{\gls{CGH}\xspace}
\newabbreviation{CS}{CS}{Compressive Sensing}

\newabbreviation{CI}{CI}{Computational Imaging}
\global\long\def\CI{\gls{CI}\xspace}
\newabbreviation{ISP}{ISP}{Image Signal Processing}
\global\long\def\ISP{\gls{ISP}\xspace}
\newabbreviation{HDR}{HDR}{High Dynamic Range}

\newabbreviation{ToF}{ToF}{Time of Flight}
\global\long\def\ToF{\gls{ToF}\xspace}

\newcolumntype{Y}{>{\centering\arraybackslash}X}
\newcolumntype{L}{>{\raggedright\arraybackslash}X}

\definecolor{Red}{rgb}{1.0, 0.0, 0.0}

\title{Does AI Understand Imaging? A Systematic Benchmark of Agentic AI for Computational Imaging Tasks}

\author{Ethan Chung$^{1,*}$, Chuanjun Zheng$^{1,*}$, Jasper Tan$^{2}$, Jingxi Li$^{2}$, Haopeng Zhang$^{1}$, and Huaijin Chen$^{1}$%
\IEEEcompsocitemizethanks{\IEEEcompsocthanksitem $^{*}$Equal contribution.\protect\\
$^{1}$University of Hawaii at Manoa.\protect\\
$^{2}$Glass Imaging.\IEEEcompsocthanksitem Preprint / work in progress. Subject to revision; not the final version.}
}

\begin{document}

\IEEEtitleabstractindextext{%
\begin{abstract}
Vision--language models (VLMs) and agentic AI have shown strong performance on semantic visual tasks, but it remains unclear whether they can handle the physics and inverse problems that underlie computational imaging. We present \textbf{ImagingBench}, a benchmark of 20 computational imaging tasks spanning five categories: ray and wave optics, image signal processing, inverse reconstruction, computational sensing, and calibration. ImagingBench evaluates three complementary settings: \textbf{Expert}, fixed expert-guided inverse reconstruction; \textbf{Planner}, planner-guided inverse reconstruction; and \textbf{Forward}, forward-system simulation for consistency checking. We benchmark leading proprietary and open-source image-centric multimodal systems, including Gemini, GPT, and Qwen, and compare them with representative task-specific non-agentic baselines. Across tasks, agentic models remain consistently weaker than specialized methods, especially on computational sensing problems such as lensless imaging, event-based reconstruction, time-of-flight imaging, and holography. Planner guidance provides only modest and inconsistent gains over the fixed-prompt Expert baseline. Although the models often generate visually plausible outputs, their reference-based fidelity remains poor, revealing a substantial gap between semantic visual competence and physically grounded imaging performance. ImagingBench provides a unified testbed for measuring this gap and tracking progress in agentic AI for computational imaging.
\end{abstract}

}

\ifpeerreview
\linenumbers \linenumbersep 15pt\relax 
\fi
\maketitle
\IEEEdisplaynontitleabstractindextext
\IEEEpeerreviewmaketitle

\begin{figure*}[t]
  \centering
  \includegraphics[width=1\linewidth]{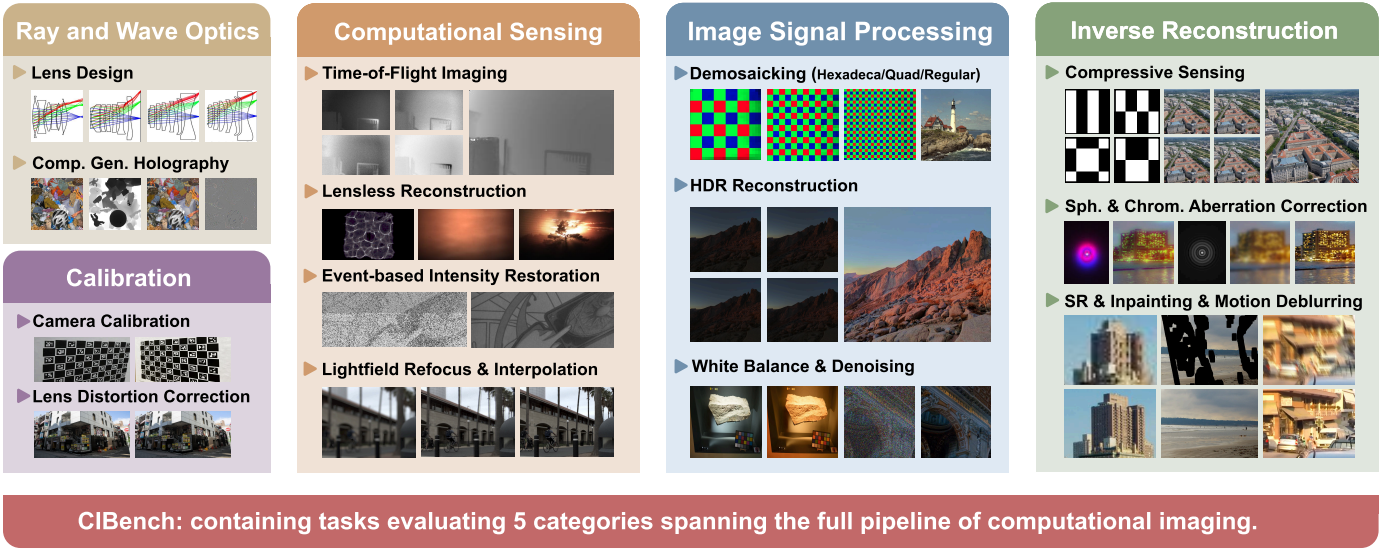}
  \caption{\textbf{Overview of ImagingBench:} Our benchmark comprises 20 subtasks spanning five computational imaging categories. Each panel shows representative tasks in the corresponding category, illustrating how the benchmark spans the computational imaging pipeline from scene acquisition and optical encoding to sensing, image processing, reconstruction, and calibration.} 
  \label{fig:system_figure}
\end{figure*}

\IEEEraisesectionheading{\section{Introduction}\label{sec:introduction}}
\IEEEPARstart{B}{enchmarks} have repeatedly reshaped progress in machine learning by turning broad capabilities into reproducible evaluation programs. In vision and multimodal reasoning, widely used suites now measure recognition, segmentation, captioning, visual question answering, tool use, physical reasoning, and controllable generation~\cite{russakovsky2015imagenet,everingham2010pascal,lin2014coco,wang2018glue,srivastava2023bigbench,liu2024mmbench,yue2024mmmu,liu2024agentbench,mialon2024gaia,chow2025physbench,yao2025mmmg}. Yet these benchmarks do not directly test a central form of visual competence: whether agentic AI systems can reason about image formation and solve computational imaging problems.

Computational imaging differs from semantic vision because success depends on optics, sensing, sampling, calibration, degradation, and reconstruction. Tasks such as denoising, deblurring, demosaicking, HDR recovery, lensless reconstruction, time-of-flight restoration, holography, calibration, and imaging-system design require models to reason about forward operators and solve ill-posed inverse problems, rather than only assign semantic labels. Classical and learning-based computational imaging has made substantial progress through physics-based modeling, task-specific optimization, and learned reconstruction pipelines~\cite{suo2023computational,wang2022differentiable,yang2024curriculum,richardson1972,lucy1974,rudin1992,zhang2017beyond,zhang2018learning}, but benchmark design has not kept pace with the emergence of multimodal foundation models and agents.

This gap matters because visually plausible outputs can be wrong when performance is governed by sensor physics, optical aberrations, noise statistics, or undersampling. It also matters for agentic systems: a useful imaging agent should not only produce an image, but also diagnose degradations, propose physically sensible restoration strategies, and remain consistent with the forward imaging process. Existing general-vision, medical-imaging, multimodal-reasoning, and agent benchmarks do not jointly evaluate these abilities in a unified setting.

To address this gap, we introduce \textbf{\ImagingBench}, a benchmark for evaluating agentic AI systems across the computational imaging pipeline. ImagingBench spans image signal processing, inverse reconstruction, computational sensing, optics-aware reasoning, and calibration, and evaluates direct execution, planner-guided execution, and forward-system simulation under a unified protocol. Our goal is not to replace broad capability suites or task-specific leaderboards, but to provide a specialized diagnostic benchmark for measuring whether modern multimodal agents exhibit coherent and transferable imaging competence.
The main contributions of this work are:
\begin{itemize}
    \item We introduce a unified benchmark for imaging-AI competence that spans 20 task groups across five categories, covering both reconstruction fidelity and physics-aware reasoning.
    \item We define three complementary evaluation settings---Expert (expert-guided inversion), Planner (planner-guided inversion), and Forward (forward-consistency evaluation)---that disentangle execution ability, per-instance planning, and forward-model consistency.
    \item We formalize task-specific evaluation using both raw metrics and normalized aggregate scores.
    \item We benchmark proprietary systems (Gemini and GPT) and an open-source system (Qwen), together with representative task-specific non-agentic baselines, and analyze where foundation-model competence transfers or fails under imaging-specific constraints.
    \item We provide cost, robustness, and failure analyses, including noise sweeps, sampling sweeps, safety-event logging, and agent-failure cases, to support reproducible future comparisons.
\end{itemize}

\begin{table*}[t]
\centering
\caption{Positioning of \textbf{ImagingBench} relative to prior benchmark families.}
\label{tab:related_work_positioning}
\renewcommand{\arraystretch}{1.15}
\resizebox{\textwidth}{!}{
\begin{tabular}{>{\raggedright\arraybackslash}m{2.8cm} >{\raggedright\arraybackslash}m{5.2cm} >{\raggedright\arraybackslash}m{4.6cm} >{\raggedright\arraybackslash}m{5.4cm}}
\toprule
\textbf{Benchmark family} & \textbf{Representative resources} & \textbf{What they evaluate well} & \textbf{What they miss for this paper} \\
\midrule
General vision &
ImageNet~\cite{russakovsky2015imagenet}, PASCAL VOC~\cite{everingham2010pascal}, COCO~\cite{lin2014coco}, VQA~\cite{antol2015vqa}, nocaps~\cite{agrawal2019nocaps}
&
recognition, localization, segmentation, captioning, visual QA
&
little or no image-formation physics, inverse problems, calibration/design, planner--executor evaluation
\\
\midrule
Medical imaging &
MedMNIST~\cite{yang2023medmnistv2}, CheXpert~\cite{irvin2019chexpert}, MIMIC-CXR~\cite{johnson2019mimiccxr}, VinDr-CXR~\cite{nguyen2022vindrcxr}, BraTS~\cite{menze2015brats}, VQA-RAD~\cite{lau2018vqarad}, SLAKE~\cite{liu2021slake}, PathVQA~\cite{he2020pathvqa}, PMC-VQA~\cite{zhang2023pmcvqa}, GMAI-MMBench~\cite{chen2024gmai_mmbench}
&
diagnosis, report-aligned learning, segmentation, lesion localization, medical VQA
&
mostly semantic/clinical understanding; limited forward-model reasoning and low-level imaging evaluation
\\
\midrule
Multimodal reasoning &
GLUE~\cite{wang2018glue}, BIG-bench~\cite{srivastava2023bigbench}, MMBench~\cite{liu2024mmbench}, MMMU~\cite{yue2024mmmu}
&
broad capability profiling, reasoning breadth, cross-task comparison
&
not imaging-specific; limited quantitative image reconstruction evaluation
\\
\midrule
Agents and physical reasoning &
AgentBench~\cite{liu2024agentbench}, GAIA~\cite{mialon2024gaia}, PhysBench~\cite{chow2025physbench}
&
tool use, multi-step planning, physical understanding
&
not end-to-end imaging pipelines; little task-specific image fidelity scoring
\\
\midrule
Multimodal generation &
MMMG~\cite{yao2025mmmg}
&
generation controllability and human-aligned auto-evaluation
&
not focused on inverse imaging or imaging-system reasoning
\\
\midrule
This work &
\textbf{ImagingBench}
&
\textbf{physical image formation, inverse reconstruction, calibration/design, and planner--executor evaluation in one framework}
&
\textbf{---}
\\
\bottomrule
\end{tabular}
}
\end{table*}
\section{Related Work}
Benchmarking has repeatedly driven progress by standardizing evaluation across broad capability families. Vision benchmarks such as ImageNet, PASCAL VOC, COCO, VQA, and nocaps established common protocols for semantic perception and visual question answering~\cite{russakovsky2015imagenet,everingham2010pascal,lin2014coco,antol2015vqa,agrawal2019nocaps}; medical-imaging benchmarks extended evaluation toward clinical interpretation~\cite{yang2023medmnistv2,irvin2019chexpert,johnson2019mimiccxr,nguyen2022vindrcxr,menze2015brats,lau2018vqarad,liu2021slake,he2020pathvqa,zhang2023pmcvqa,chen2024gmai_mmbench}; and recent multimodal and agentic suites evaluate reasoning, tool use, physical understanding, and controllable generation~\cite{wang2018glue,srivastava2023bigbench,liu2024mmbench,yue2024mmmu,liu2024agentbench,mialon2024gaia,chow2025physbench,luo2025mmmg}. However, these benchmark families still leave an important gap for computational imaging. Our setting requires models to reason jointly about image formation, optics, sensing, inverse reconstruction, calibration, and agentic planning. High performance on semantic recognition or multimodal QA does not imply competence on physically grounded imaging tasks, where success depends on correctly handling forward operators, degradations, and reconstruction objectives.
\subsection{Computational Imaging, Physics-Aware Restoration, and Optical Design}

Classical and learning-based \CI has advanced primarily through physics-grounded forward modeling, inverse methods, and task-specific neural architectures~\cite{suo2023computational,wang2022differentiable,yang2024curriculum,chang2018hybrid}. Recent work has also begun to explore how foundation-model priors can assist low-level imaging. For example, VLMIR injects vision-language priors into a diffusion restoration pipeline through VLM-based feature extraction and cross-attention conditioning~\cite{yang2025vlmir}. In optical engineering, OPTIAGENT introduces an agentic framework for automated optical design, combining an LLM, a curated optical-design dataset, physics-driven rewards, and downstream optimization routines~\cite{geng2026optiagent}. These works are important steps toward integrating multimodal foundation models with imaging tasks, but they remain method-centric and confined to relatively narrow task families. They do not provide a unified benchmark spanning image formation, inverse recovery, calibration, and planner--executor evaluation.
\subsection{General Vision and Medical Imaging Benchmarks}

General vision benchmarks have been highly successful at standardizing evaluation for semantic perception tasks. ImageNet emphasized large-scale classification and detection; PASCAL VOC standardized recognition and localization; COCO expanded toward instance segmentation and captioning in natural scenes; VQA added language-conditioned reasoning over images; and nocaps stressed captioning under novel-object generalization~\cite{russakovsky2015imagenet,everingham2010pascal,lin2014coco,antol2015vqa,agrawal2019nocaps}. These resources evaluate semantic understanding well, but they provide little or no coverage of forward image formation, inverse imaging, calibration, or design.

Medical imaging benchmarks similarly focus on clinically meaningful interpretation. MedMNIST provides a lightweight multi-dataset biomedical classification suite; CheXpert, MIMIC-CXR, and VinDr-CXR support chest X-ray classification and localization; BraTS is a major benchmark for tumor segmentation; and VQA-RAD, SLAKE, PathVQA, PMC-VQA, and GMAI-MMBench extend evaluation to medical multimodal reasoning and visual question answering~\cite{yang2023medmnistv2,irvin2019chexpert,johnson2019mimiccxr,nguyen2022vindrcxr,menze2015brats,lau2018vqarad,liu2021slake,he2020pathvqa,zhang2023pmcvqa,chen2024gmai_mmbench}. Yet these benchmarks are still mostly centered on semantic or clinical interpretation, rather than on reasoning about sensing operators, optical degradation, inverse reconstruction fidelity, or physically consistent forward simulation.
\subsection{Multimodal Reasoning, Agents, and Physical Understanding}

A parallel line of work studies increasingly general multimodal or agentic capability. GLUE and BIG-bench emphasize broad cross-task evaluation in language and reasoning~\cite{wang2018glue,srivastava2023bigbench}. MMBench and MMMU probe the breadth of multimodal reasoning and perception~\cite{liu2024mmbench,yue2024mmmu}. AgentBench and GAIA evaluate LLMs and multimodal systems as agents in interactive or tool-using settings~\cite{liu2024agentbench,mialon2024gaia}. PhysBench focuses specifically on physical-world understanding, showing that current VLMs still struggle with many physically grounded scenarios~\cite{chow2025physbench}. MMMG studies controllable multimodal generation with human-aligned automatic evaluation across multiple modality combinations~\cite{yao2025mmmg}. 
Complementarily, Luo et al.\ introduce a text-to-image reasoning benchmark centered on knowledge-image generation across disciplines and educational levels~\cite{luo2025mmmg}. These efforts are highly relevant to our study, especially for planner--executor design and physical reasoning, but they are not imaging-specific and do not evaluate quantitative inverse reconstruction quality across a diverse computational imaging pipeline.
\subsection{How Our Benchmark Differs}

Table~\ref{tab:related_work_positioning} summarizes the distinction between prior benchmark families and our setting. Relative to general vision benchmarks, our benchmark evaluates physical image formation and inverse recovery rather than only semantic perception. Relative to medical imaging benchmarks, it emphasizes low-level imaging physics in addition to multimodal reasoning. Relative to multimodal reasoning and agent benchmarks, it provides quantitative image-fidelity evaluation and forward--inverse consistency checks. Relative to generation benchmarks, it focuses on physically grounded imaging tasks rather than free-form image generation alone.

ImagingBench is intended as a specialized diagnostic benchmark rather than a replacement for broad capability suites or task-specific leaderboards. Its purpose is to isolate whether frontier multimodal agents can transfer into operator-aware computational imaging under a unified, reproducible evaluation protocol.

To the best of our knowledge, \textbf{ImagingBench} is the first benchmark to jointly evaluate physical image formation, inverse reconstruction, calibration/design, and planner--executor behavior in a single computational imaging framework.

\section{Benchmark Construction}\label{sec:benchmark}

\subsection{Evaluated Tasks}
\begin{table*}
\caption{Summary of benchmark subtasks, task descriptions, input/output modalities, and evaluation metrics.}
\label{tab:task_summary}
\centering
\scriptsize
\renewcommand{\arraystretch}{1.2}
\setlength{\tabcolsep}{3pt}

\begin{tabularx}{\textwidth}{l L ccc}
\toprule
\textbf{Task} & \textbf{Description} & \textbf{In $\to$ Out} & \textbf{Inst.} & \textbf{Eval.} \\
\midrule

\multicolumn{5}{l}{\cellcolor[gray]{0.9}\textbf{Ray and Wave Optics}} \\
Lens System Design & Design a multi-element optical lens system from textual specifications. & T $\to$ T & --- & RMS \\
Comp.\ Gen.\ Holography & Reconstruct an image from its computer-generated hologram (amplitude and phase). & T, \imgicon{example-image} $\to$ \imgicon{example-image} & 100 & P/S \\
\hline

\multicolumn{5}{l}{\cellcolor[gray]{0.9}\textbf{Image Signal Processing}} \\
Demosaicking (Hexadeca) & Reconstruct a full-color image from a hexadeca ($8{\times}8$ super-pixel) Bayer mosaic. & T, \imgicon{example-image} $\to$ \imgicon{example-image} & 100 & P/S \\
Demosaicking (Quad) & Reconstruct a full-color image from a quad ($4{\times}4$ super-pixel) Bayer mosaic. & T, \imgicon{example-image} $\to$ \imgicon{example-image} & 100 & P/S \\
Demosaicking (Regular) & Reconstruct a full-color image from a quad ($2{\times}2$ super-pixel) Bayer mosaic. & T, \imgicon{example-image} $\to$ \imgicon{example-image} & 100 & P/S \\
High Dynamic Range & Recover a well-exposed image from underexposed raw burst frames. & T, \imgicon{example-image} $\to$ \imgicon{example-image} & 100 & P/S \\
White Balance & Correct the color cast of an image captured under a non-neutral illuminant. & T, \imgicon{example-image} $\to$ \imgicon{example-image} & 100 & P/S \\
Denoising & Remove real sensor noise from smartphone camera images. & T, \imgicon{example-image} $\to$ \imgicon{example-image} & 100 & P/S/L \\
\hline

\multicolumn{5}{l}{\cellcolor[gray]{0.9}\textbf{Inverse Reconstruction}} \\
Spherical Aberration & Remove blur caused by spherical aberration at varying wavefront-error levels. & T, \imgicon{example-image} $\to$ \imgicon{example-image} & 100 & P/S \\
Longitudinal Chromatic & Correct per-channel defocus from wavelength-dependent focal-length shifts. & T, \imgicon{example-image} $\to$ \imgicon{example-image} & 100 & P/S \\
Motion Blur & Recover a sharp image from real camera or object motion blur. & T, \imgicon{example-image} $\to$ \imgicon{example-image} & 100 & P/S \\
Randomly Masked Sub-Sampling & Reconstruct a full image from randomly sampled pixels at various sampling ratios. & T, \imgicon{example-image} $\to$ \imgicon{example-image} & 100 & P/S \\
Comp.\ Sensing (block) & Reconstruct a full image from block-wise compressed measurements at various ratios. & T, \imgicon{example-image} $\to$ \imgicon{example-image} & 100 & P/S \\
Super-Resolution & Recover a high-resolution image from a downsampled low-resolution input. & T, \imgicon{example-image} $\to$ \imgicon{example-image} & 100 & P/S \\
Inpainting & Fill in masked (missing) image regions with plausible content. & T, \imgicon{example-image} $\to$ \imgicon{example-image} & 100 & P/S \\
\hline

\multicolumn{5}{l}{\cellcolor[gray]{0.9}\textbf{Computational Sensing}} \\
Lensless Imaging & Recover the scene from a lensless camera measurement. & T, \imgicon{example-image} $\to$ \imgicon{example-image} & 100 & P/S \\
Lightfield View Extrapolation & Synthesize an extreme off-center view given clustered sub-aperture center views. & T, \imgicon{example-image} $\to$ \imgicon{example-image} & 100 & P/S/L \\
Event-based Intensity Imaging & Reconstruct an intensity image from event streams. & T, \imgicon{example-image} $\to$ \imgicon{example-image} & 100 & P/S \\
Time-of-Flight Depth Imaging & Reconstruct a depth map from raw time-of-flight sensor measurements. & T, \imgicon{example-image} $\to$ \imgicon{example-image} & 100 & P/S \\
\hline

\multicolumn{5}{l}{\cellcolor[gray]{0.9}\textbf{Calibration}} \\
Camera Calibration & Estimate intrinsic camera parameters from checkerboard calibration images. & T, \imgicon{example-image} $\to$ T & --- & P/S \\

\bottomrule
\end{tabularx}
\vspace{2pt}
{\raggedright\footnotesize
T~=~text prompt; \imgicon{example-image}~=~image; Inst.~=~number of test instances.
P/S~=~PSNR/SSIM; P/S/L~=~PSNR/SSIM/LPIPS.\par}
\end{table*}

\CI extends conventional digital imaging by jointly designing image formation and reconstruction. Rather than treating image formation as a fixed front-end followed by generic post-processing, \CI integrates physics-driven acquisition (optics and sensors) with computational reconstruction and inference~\cite{suo2023computational}. 
In \ImagingBench, we evaluate whether \VLMs can meaningfully assist \CI workflows. The benchmark comprises 20 subtasks organized into five categories (Fig.~\ref{fig:system_figure}), covering the computational imaging pipeline from optical design and computational sensing to image signal processing, inverse reconstruction, and calibration. The categories are summarized below and detailed in Table~\ref{tab:task_summary}:

\begin{itemize}
    \item \textbf{Ray and Wave Optics:} ray-optics-based lens design and \CGH~\cite{wang2022differentiable,yang2024curriculum,shi2021towards}.
    \item \textbf{Computational sensing:} event-camera intensity image reconstruction~\cite{gao2025unified}, \ToF imaging~\cite{qiao2022depth}, lensless reconstruction~\cite{monakhova2019learned}, and \CGH~\cite{shi2021towards}.
    \item \textbf{Image signal processing:} white balance~\cite{afifi2019color}, demosaicking~\cite{a2021beyond}, denoising~\cite{zhang2017beyond}, and HDR reconstruction~\cite{hdrplus_hasinoff2016burst}.
    \item \textbf{Inverse reconstruction:} deconvolution (optical-blur removal such as spherical/chromatic aberrations) and undersampled reconstruction (compressive sensing~\cite{chen2015fpa}, super-resolution~\cite{zhang2018learning}, and inpainting~\cite{suvorov2022resolution}).
    \item \textbf{Calibration:} intrinsic camera-parameter estimation from checkerboard calibration images.
\end{itemize}

\begin{figure*}[t]
  \centering
  \includegraphics[width=1.0\linewidth]{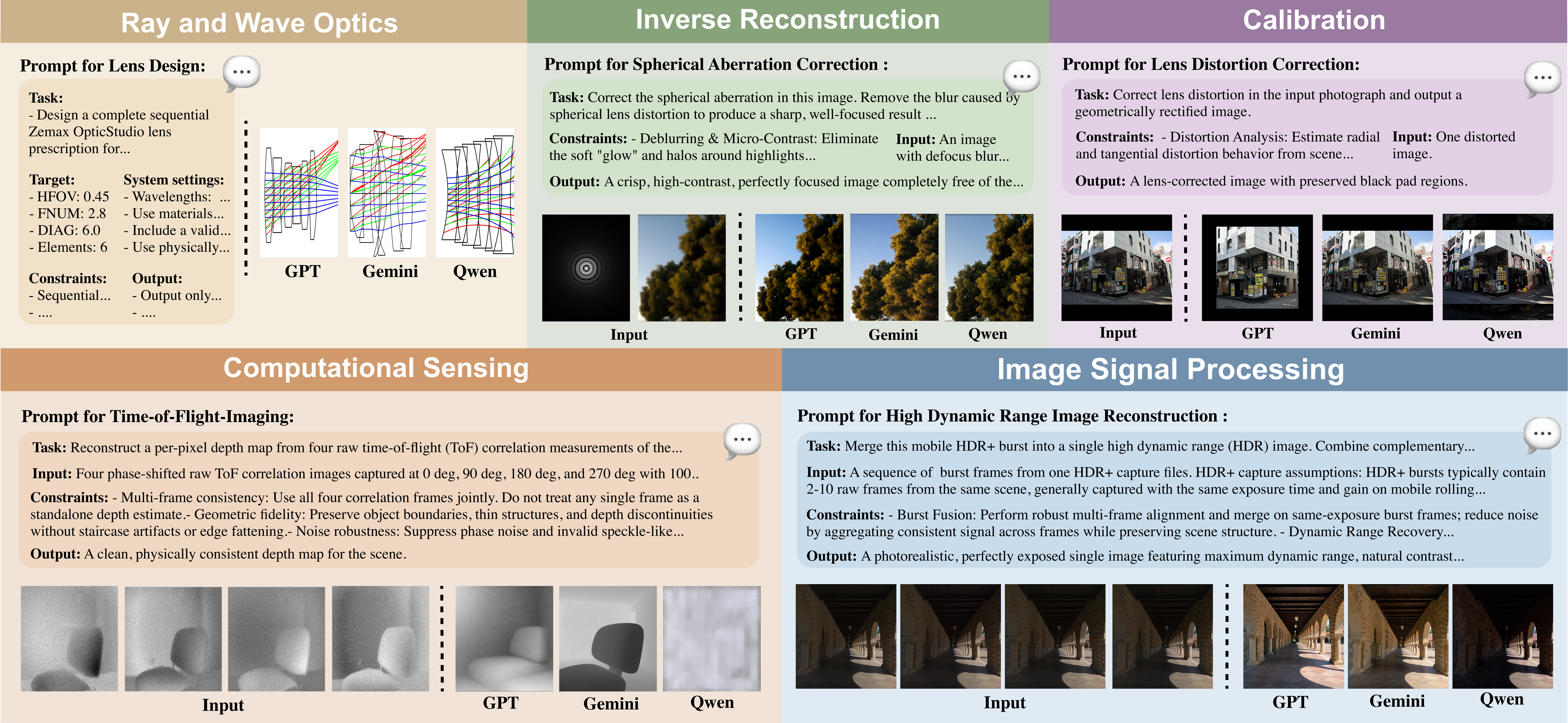}
  \caption{\textbf{Representative qualitative examples from ImagingBench}: 
  For each example, we show the task prompt and representative outputs produced by three frontier agentic AI systems (GPT, Gemini, and Qwen), highlighting the breadth of ImagingBench across physically grounded inverse problems and structured imaging-reasoning tasks. The examples also illustrate substantial variation in output fidelity, physical plausibility, and task adherence across models.
  }
  \label{fig:examples}
\end{figure*}

\noindent For the evaluation, each task family $t$ can be represented as
\begin{equation}
    \mathcal{D}_t=\{(x_i,q_i,y_i,m_i)\}_{i=1}^{N_t},
    \label{eq:tasks}
\end{equation}
where $x_i$ is the model input, $q_i$ is an optional text instruction, $y_i$ is the target output, and $m_i$ contains task-specific metadata such as sensor parameters, exposure settings, wavelength, sampling ratio, or calibration geometry. To avoid overloading $x_i$, we use $z_i$ to denote the latent scene, clean image, or underlying physical quantity before measurement. The benchmark input $x_i$ is generated from $z_i$ by a task-dependent forward model, while $y_i$ is the desired reconstruction, inferred parameter, or design output. Fig.~\ref{fig:examples} shows representative tasks and outputs; the supplementary material provides the detailed formulation for each task.

\subsection{Evaluation Protocol}

Building on Eq.~\ref{eq:tasks}, a model $f_t^{(k)}$ with parameters $\theta_k$ produces
\[
\hat{y}_i = f_t^{(k)}(x_i, q_i, m_i; \theta_k).
\]

This notation covers both inverse and forward computational imaging tasks. For inverse problems, $x_i$ is typically a degraded measurement and $y_i$ is the desired reconstruction. For forward tasks, $x_i$ may instead be a clean image or latent signal, and $y_i$ is the corresponding simulated measurement under a known image-formation model. For agentic systems, the mapping from input to output may also be mediated by a planner interacting with tools over multiple steps. We write this planner--executor process as
\[
a_\tau \sim \pi_\phi(a_\tau \mid s_\tau), \quad
s_{\tau+1} = T(s_\tau, a_\tau), \quad
\hat{y} = g_\psi(s_{0:K}),
\]
where $\pi_\phi$ is the planner policy, $T$ is the tool or environment transition, and $g_\psi$ is the final executor or answer generator.

\begin{figure*}
    \centering
    \includegraphics[width=1.0\linewidth]{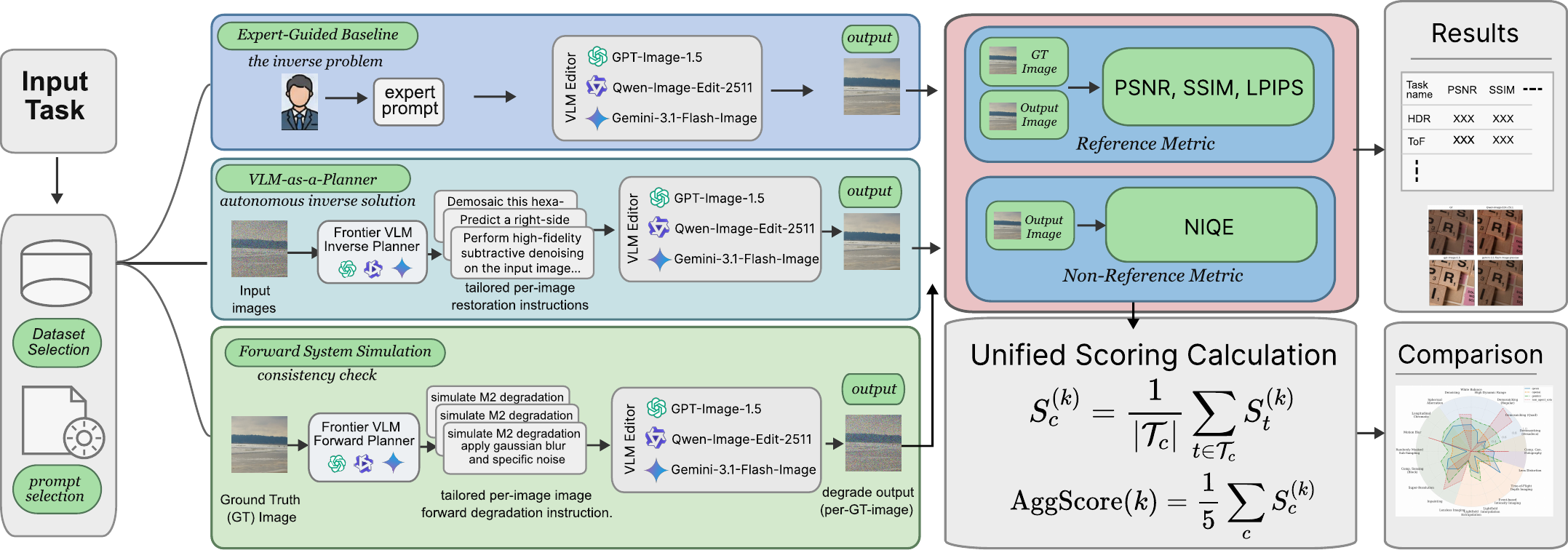}
    \caption{\textbf{Overview of the ImagingBench evaluation pipeline:} Each input task is paired with a selected dataset and prompt template, then evaluated under three protocols: the Expert protocol (expert-guided inverse reconstruction), the Planner protocol (planner-guided inverse reconstruction), and the Forward protocol (forward-system simulation). In Planner and Forward, a frontier agentic AI planner generates image-specific instructions that are executed by a VLM image editor. Outputs are scored using reference metrics (PSNR, SSIM, LPIPS) and the no-reference metric NIQE, then aggregated into a unified category-level score for final comparison across models and protocols.}
    \label{fig:evaluation_pipeline}
\end{figure*}

The evaluation pipeline is illustrated in Fig.~\ref{fig:evaluation_pipeline}. To probe different aspects of imaging competence, we define three complementary paradigms based on the computational imaging formalism
\[
x = \mathcal{A}(z; m) + n,
\]
where $z$ denotes the latent clean signal, $\mathcal{A}$ is the task-dependent forward operator, and $n$ represents noise or corruption. These paradigms are designed to isolate direct execution ability, per-image planning ability, and conceptual consistency between the forward model and its inverse.
In this paper, ``agentic'' refers specifically to the observe--plan--execute structure of the Planner protocol: the system observes the degraded input, plans an image-specific restoration instruction, and then executes it through the editor. The resulting protocols evaluate operator-conditioned behavior, but they do not by themselves prove that the model causally uses the provided forward model rather than generic image priors.

\subsubsection{Expert Protocol: Fixed-Prompt Inverse Execution}

In Expert, we evaluate the model's ability to solve the inverse problem under a fixed, domain-informed expert instruction. For inverse tasks, this corresponds to recovering the desired target $y_i$ from the degraded input $x_i$ using a shared prompt $q^{\mathrm{exp}}$ across all samples in the task:
\[
\hat{y}_i = f_t^{(k)}(x_i, q^{\mathrm{exp}}, m_i; \theta_k).
\]
This setting serves as a standardized baseline for execution quality. The expert prompt encodes task-relevant restoration guidance while imposing constraints intended to reduce structural hallucination, artificial texture synthesis, and other undesired generative artifacts. As such, Expert isolates the core execution capability of the underlying image-editing or multimodal model under strong but fixed human supervision.

\subsubsection{Planner Protocol: Adaptive Inverse Solution}

In Planner, we replace the fixed expert instruction with a per-image adaptive plan generated by a VLM planner. The planner first analyzes the degraded input and associated metadata, then produces a tailored restoration prompt for the executor:
\[
q_i^{\mathrm{plan}} = \pi_\phi(x_i, m_i), \qquad
\hat{y}_i = f_t^{(k)}(x_i, q_i^{\mathrm{plan}}, m_i; \theta_k).
\]
In practice, the planner receives the input image together with a minimal instruction such as ``Analyze and restore this,'' and outputs an image-specific prompt describing the likely degradation and a suitable restoration strategy. This prompt is then passed to the corresponding editing model. By comparing Planner against Expert, we test whether adaptive multimodal planning provides a measurable prompting gain over a strong generalized expert prompt.

\subsubsection{Forward Protocol: System Simulation (Consistency Check)}

While Expert and Planner evaluate inverse execution, Forward probes whether the model exhibits a coherent understanding of the forward image-formation process itself. In this setting, the model is given a clean signal or image and asked to synthesize the degraded measurement predicted by the task-specific forward model:
\[
\hat{y}_i = f_t^{(k)}(z_i, q_i, m_i; \theta_k),
\]
where the target corresponds to the known forward output associated with the task. For example, the model may be asked to add stochastic noise, optical blur, or undersampling artifacts consistent with the degradations encountered in Expert and Planner.

Conceptually, Forward reverses the direction of the inverse tasks: rather than removing degradation, the model must generate it in a physically meaningful way. Strong performance in this setting indicates not only pattern-matching ability, but also consistency with the underlying sensing or imaging process.

For reproducibility, the supplementary material includes representative Expert/Planner/Forward prompts together with paired input/output examples.

\subsection{Evaluation Metrics}

For each task family $t$ and sample $i$, the model produces a prediction
$\hat{y}_i = f_t^{(k)}(x_i,q_i,m_i;\theta_k)$, which is compared against the task target $y_i$ using task-appropriate metrics. In line with the protocol above, we report \emph{raw task metrics first}, and use a normalized aggregate score only as a secondary summary across heterogeneous tasks.

\vspace{0.5em}
\noindent\textbf{Reconstruction-based tasks.}
For image reconstruction and restoration tasks, both $y_i$ and $\hat{y}_i$ are images defined on the same valid support $\Omega_i$ 
Let $|\Omega_i|$ denote the number of valid pixels, and let the dynamic range be $L$ (we use $L=1$ for normalized images). We evaluate four complementary metrics: PSNR, SSIM, LPIPS, and NIQE. 

\vspace{0.5em}
\noindent\textbf{Normalized reconstruction score.}
Because these metrics have different scales and directions, we additionally report a normalized per-instance reconstruction score for cross-task summarization. For sample $i$, we define
\begin{align}
s^{\mathrm{psnr}}_i
&=
\mathrm{clip}\!\left(
\frac{\mathrm{PSNR}(y_i,\hat{y}_i)-15}{40-15},\,0,\,1
\right), \\
s^{\mathrm{ssim}}_i
&=
\mathrm{clip}\!\left(
\mathrm{SSIM}(y_i,\hat{y}_i),\,0,\,1
\right), \\
s^{\mathrm{lpips}}_i
&=
\mathrm{clip}\!\left(
1-\mathrm{LPIPS}(y_i,\hat{y}_i),\,0,\,1
\right), \\
s^{\mathrm{niqe}}_i
&=
\mathrm{clip}\!\left(
\frac{20-\mathrm{NIQE}(\hat{y}_i)}{17},\,0,\,1
\right).
\end{align}
We then combine them into an instance-level weighted score
\[
s_i^{\mathrm{recon}}
=
\frac{
w_{\mathrm{psnr}} s_i^{\mathrm{psnr}}
+
w_{\mathrm{ssim}} s_i^{\mathrm{ssim}}
+
w_{\mathrm{lpips}} s_i^{\mathrm{lpips}}
+
w_{\mathrm{niqe}} s_i^{\mathrm{niqe}}
}{
w_{\mathrm{psnr}}+w_{\mathrm{ssim}}+w_{\mathrm{lpips}}+w_{\mathrm{niqe}}
},
\]
with base weights
\[
w_{\mathrm{psnr}}=0.3,\quad
w_{\mathrm{ssim}}=0.3,\quad
w_{\mathrm{lpips}}=0.3,\quad
w_{\mathrm{niqe}}=0.1.
\]
We weight PSNR, SSIM, and LPIPS equally because they provide complementary reference-based fidelity views, while NIQE receives a smaller weight because it is no-reference and is intended to inform the aggregate without dominating it.
If a metric is unavailable for a particular sample, its weight is set to zero and the remaining weights are renormalized.

For each reconstruction task $t$, we report both the raw metric means and the averaged normalized score
\[
S_t^{\mathrm{recon}}
=
\frac{1}{N_t}\sum_{i=1}^{N_t} s_i^{\mathrm{recon}}.
\]

\vspace{0.5em}
\noindent\textbf{Task-specific metrics beyond reconstruction.}
Not all tasks in \ImagingBench are image-to-image restoration problems. 
For example, lens-design tasks are evaluated by optical validity and downstream image-quality criteria such as spot size, MTF, and Strehl ratio, while forward-simulation tasks are evaluated by consistency with the known image-formation process. We denote such task-specific scores generically by
\[
\mathrm{Metric}_t\!\left(y_i,\hat{y}_i;m_i\right),
\]
to emphasize that the exact metric depends on the task.

\vspace{0.5em}
\noindent\textbf{Aggregate score across tasks.}
To summarize performance across heterogeneous task families, we use a two-level
macro-average. Each task contributes a task-level score
\[
S_t^{(k)} =
\begin{cases}
S_t^{\mathrm{recon},(k)}, & \text{reconstruction-based task},\\[0.4em]
\mathrm{Norm}_t\!\left(\mathrm{Metric}_t^{(k)}\right), & \text{otherwise},
\end{cases}
\]
where $S_t^{\mathrm{recon},(k)}$ is the averaged normalized reconstruction score
and $\mathrm{Norm}_t(\cdot)$ maps a non-reconstruction metric to a common
higher-is-better $[0,1]$ scale. Camera calibration uses
$\mathrm{Norm}_t=\mathrm{clip}(1-\mathrm{MRE},0,1)$ over its intrinsic and
distortion relative errors (parse failures score $0$), and lens design uses
$\mathrm{Norm}_t=\mathrm{clip}(\mathrm{Strehl}/0.8,0,1)$ against the
diffraction-limited threshold (unanalyzable targets score $0$). We then average
within each of the five task categories $c$ and across categories:
\[
S_c^{(k)} = \frac{1}{|\mathcal{T}_c|}\sum_{t\in\mathcal{T}_c} S_t^{(k)},
\qquad
\mathrm{AggScore}(k) = \frac{1}{5}\sum_{c} S_c^{(k)}.
\]
Calibration and lens design enter the Calibration and Ray-and-Wave-Optics
categories, respectively. We use a category macro-average rather than a flat
per-task average so that families with many subtasks (e.g.\ image signal
processing) do not dominate the score; the model ranking is unchanged under flat
$1/|\mathcal{T}|$ weighting (supplement). The aggregate score is used only for
cross-task summarization; the primary analysis remains the raw task metrics
reported for each subtask.

\subsection{Evaluated Models}

We evaluate representative state-of-the-art image-centric multimodal systems spanning both proprietary and open-source families. For our base editing execution (Expert), we utilize \textbf{Nano Banana 2} (Gemini 3.1 Flash Image Preview) \cite{google2026gemini31flash}, \textbf{GPT-Image-1.5} \cite{openai2026gptimage15}, and \textbf{Qwen-Image-Edit-2511} \cite{wu2025qwenimagetechnicalreport}. Nano Banana 2 is accessed via Google Vertex AI (checkpoint: 2026-02-26), and GPT-Image-1.5 is accessed via Azure Foundry (version: 2025-12-16). Qwen-Image-Edit-2511 is hosted on 1xNVIDIA H200 GPU and run with 28-step diffusion.

To support the planner-guided setting (Planner), we pair these base editing models with their respective frontier multimodal reasoning models (acting as the planner): \textbf{Gemini 3.1 Pro Preview}, \textbf{GPT-5}, and \textbf{Qwen3.5-35B-A3B}.

For Gemini and GPT, we use default API inference settings. For Qwen-Image-Edit-2511, we follow the official HuggingFace implementation with native 28-step diffusion while keeping all other parameters at default values.

To reduce model-specific auto-resizing artifacts, all benchmark inputs are standardized to a $1024 \times 1024$ canvas before inference. In practice, this avoids unintended scale changes from provider-side resolution bucketing and keeps output size consistent for downstream scoring. 
For tasks where resizing would remove task-defining structure, we instead pad images to $1024 \times 1024$ and evaluate predictions only on the unpadded crop. \label{sec:model_crop}

\section{Data Collection and Quality Assurance}

As shown in Fig.~\ref{fig:dataset-task-grouped}, \ImagingBench combines existing datasets with newly constructed data. It contains three components: (1) the main benchmark, (2) an ablation suite for controlled sensitivity analysis, and (3) a hidden test set for fair evaluation on unseen examples. We summarize the main curation procedure here and defer additional preprocessing details to the supplementary material.
We target roughly 100 examples per subtask because \ImagingBench is intended as a diagnostic zero-shot benchmark spanning many computational-imaging operators rather than as a large-scale training resource.

\subsection{Main Benchmark Data Collection}

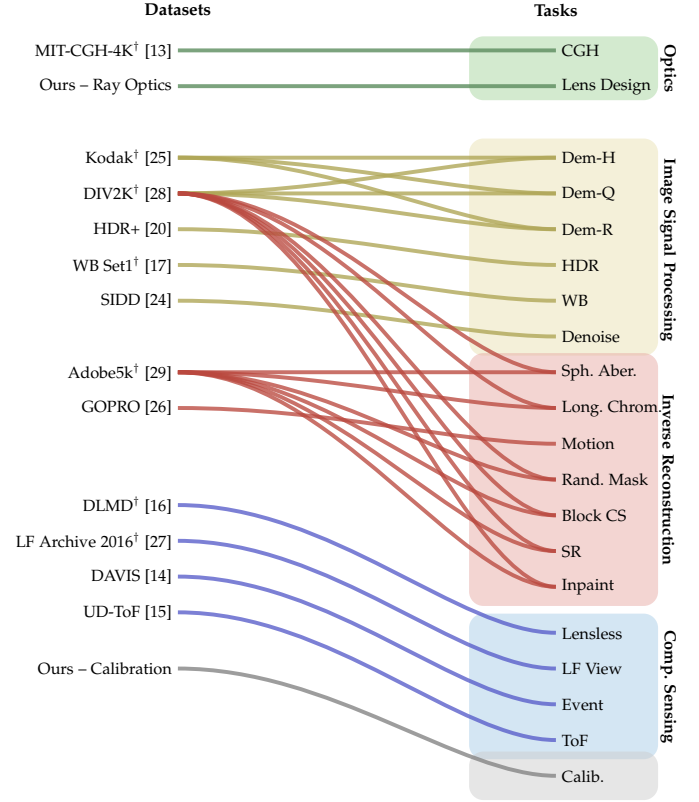
\begin{figure}[t]
\centering
\resizebox{1\columnwidth}{!}{%
\begin{tikzpicture}[
    dataset/.style={anchor=east, font=\small},
    task/.style={anchor=west, font=\small},
    catlabel/.style={font=\bfseries\small},
    catbox/.style={rounded corners=8pt, draw=none, inner sep=5pt},
    linkrw/.style={draw={rgb,255:red,92; green,140; blue,94}, line width=2.2pt, opacity=0.80},
    linkisp/.style={draw={rgb,255:red,176; green,167; blue,88}, line width=2.2pt, opacity=0.80},
    linkinv/.style={draw={rgb,255:red,186; green,73; blue,63}, line width=2.2pt, opacity=0.78},
    linkcs/.style={draw={rgb,255:red,87; green,92; blue,201}, line width=2.2pt, opacity=0.78},
    linkcal/.style={draw={rgb,255:red,130; green,130; blue,130}, line width=2.2pt, opacity=0.78}
]

\node[font=\bfseries] at (0,12.1) {Datasets};
\node[font=\bfseries] at (7.4,12.1) {Tasks};

\node[task] (cgh)   at (7.4,11.3) {CGH};
\node[task] (lensd) at (7.4,10.6) {Lens Design};

\node[task] (demh)  at (7.4,9.2)  {Dem-H};
\node[task] (demq)  at (7.4,8.5)  {Dem-Q};
\node[task] (demr)  at (7.4,7.8)  {Dem-R};
\node[task] (hdr)   at (7.4,7.1)  {HDR};
\node[task] (wbt)   at (7.4,6.4)  {WB};
\node[task] (den)   at (7.4,5.7)  {Denoise};

\node[task] (sph)   at (7.4,5.0)  {Sph.\ Aber.};
\node[task] (lca)   at (7.4,4.3)  {Long.\ Chrom.};
\node[task] (mot)   at (7.4,3.6)  {Motion};
\node[task] (rms)   at (7.4,2.9)  {Rand.\ Mask};
\node[task] (blk)   at (7.4,2.2)  {Block CS};
\node[task] (sr)    at (7.4,1.5)  {SR};
\node[task] (inp)   at (7.4,0.8)  {Inpaint};

\node[task] (lens)  at (7.4,-0.1) {Lensless};
\node[task] (lfv)   at (7.4,-0.8) {LF View};
\node[task] (evt)   at (7.4,-1.5) {Event};
\node[task] (tof)   at (7.4,-2.2) {ToF};

\node[task] (cal)   at (7.4,-2.9) {Calib.};

\begin{scope}[on background layer]
    \node[catbox, fill={rgb,255:red,168; green,214; blue,160}, opacity=0.5,
          minimum width=3.7cm, minimum height=1.25cm] (boxrw) at (7.55,10.95) {};
    \node[catbox, fill={rgb,255:red,234; green,226; blue,179}, opacity=0.5,
          minimum width=3.7cm, minimum height=4.25cm] (boxisp) at (7.55,7.45) {};
    \node[catbox, fill={rgb,255:red,231; green,171; blue,166}, opacity=0.5,
          minimum width=3.7cm, minimum height=4.95cm] (boxinv) at (7.55,2.9) {};
    \node[catbox, fill={rgb,255:red,170; green,206; blue,232}, opacity=0.5,
          minimum width=3.7cm, minimum height=2.85cm] (boxcs) at (7.55,-1.15) {};
    \node[catbox, fill={rgb,255:red,205; green,205; blue,205}, opacity=0.5,
          minimum width=3.7cm, minimum height=0.95cm] (boxcal) at (7.55,-2.9) {};
\end{scope}

\node[catlabel, rotate=-90] at (9.6,10.95) {Optics};
\node[catlabel, rotate=-90] at (9.6,7.4)   {Image Signal Processing};
\node[catlabel, rotate=-90] at (9.6,2.9)   {Inverse Reconstruction};
\node[catlabel, rotate=-90] at (9.6,-1.15)  {Comp.\ Sensing};
\node[catlabel, rotate=-90] at (9.6,-2.9)  {};
\node[dataset] (mit)      at (0,11.3) {MIT-CGH-4K$^\dagger$~[13]};
\node[dataset] (oursrw)   at (0,10.6) {Ours -- Ray Optics};

\node[dataset] (kodak)    at (0,9.2)  {Kodak$^\dagger$~[25]};
\node[dataset] (div2k)    at (0,8.5)  {DIV2K$^\dagger$~[28]};
\node[dataset] (hdrp)     at (0,7.8)  {HDR+~[20]};
\node[dataset] (wb)       at (0,7.1)  {WB Set1$^\dagger$~[17]};
\node[dataset] (sidd)     at (0,6.4)  {SIDD~[24]};

\node[dataset] (adobe)    at (0,5.0)  {Adobe5k$^\dagger$~[29]};
\node[dataset] (gopro)    at (0,4.3)  {GOPRO~[26]};

\node[dataset] (dlmd)     at (0,2.4)  {DLMD$^\dagger$~[16]};
\node[dataset] (lfa)      at (0,1.7)  {LF Archive 2016$^\dagger$~[27]};
\node[dataset] (davis)    at (0,1.0)  {DAVIS~[14]};
\node[dataset] (udtof)    at (0,0.3)  {UD-ToF~[15]};

\node[dataset] (ourscal)  at (0,-0.8) {Ours -- Calibration};

\draw[linkrw]  (mit.east)     .. controls (3.0,11.3) and (5.5,11.3) .. (cgh.west);
\draw[linkrw]  (oursrw.east)  .. controls (3.0,10.6) and (5.5,10.6) .. (lensd.west);

\draw[linkisp] (kodak.east)   .. controls (3.0,9.2) and (5.5,9.2) .. (demh.west);
\draw[linkisp] (kodak.east)   .. controls (3.0,9.2) and (5.5,8.5) .. (demq.west);
\draw[linkisp] (kodak.east)   .. controls (3.0,9.2) and (5.5,7.8) .. (demr.west);

\draw[linkisp] (div2k.east)   .. controls (3.0,8.5) and (5.5,9.2) .. (demh.west);
\draw[linkisp] (div2k.east)   .. controls (3.0,8.5) and (5.5,8.5) .. (demq.west);
\draw[linkisp] (div2k.east)   .. controls (3.0,8.5) and (5.5,7.8) .. (demr.west);

\draw[linkisp] (hdrp.east)    .. controls (3.0,7.8) and (5.5,7.1) .. (hdr.west);
\draw[linkisp] (wb.east)      .. controls (3.0,7.1) and (5.5,6.4) .. (wbt.west);
\draw[linkisp] (sidd.east)    .. controls (3.0,6.4) and (5.5,5.7) .. (den.west);

\draw[linkinv] (adobe.east)   .. controls (3.0,5.0) and (5.5,5.0) .. (sph.west);
\draw[linkinv] (adobe.east)   .. controls (3.0,5.0) and (5.5,4.3) .. (lca.west);
\draw[linkinv] (adobe.east)   .. controls (3.0,5.0) and (5.5,2.9) .. (rms.west);
\draw[linkinv] (adobe.east)   .. controls (3.0,5.0) and (5.5,2.2) .. (blk.west);
\draw[linkinv] (adobe.east)   .. controls (3.0,5.0) and (5.5,1.5) .. (sr.west);
\draw[linkinv] (adobe.east)   .. controls (3.0,5.0) and (5.5,0.8) .. (inp.west);

\draw[linkinv] (div2k.east)   .. controls (3.0,8.5) and (5.5,5.0) .. (sph.west);
\draw[linkinv] (div2k.east)   .. controls (3.0,8.5) and (5.5,4.3) .. (lca.west);
\draw[linkinv] (div2k.east)   .. controls (3.0,8.5) and (5.5,2.9) .. (rms.west);
\draw[linkinv] (div2k.east)   .. controls (3.0,8.5) and (5.5,2.2) .. (blk.west);
\draw[linkinv] (div2k.east)   .. controls (3.0,8.5) and (5.5,1.5) .. (sr.west);
\draw[linkinv] (div2k.east)   .. controls (3.0,8.5) and (5.5,0.8) .. (inp.west);

\draw[linkinv] (gopro.east)   .. controls (3.0,4.3) and (5.5,3.6) .. (mot.west);

\draw[linkcs]  (dlmd.east)    .. controls (3.0,2.4) and (5.5,-0.1) .. (lens.west);
\draw[linkcs]  (lfa.east)     .. controls (3.0,1.7) and (5.5,-0.8) .. (lfv.west);
\draw[linkcs]  (davis.east)   .. controls (3.0,1.0) and (5.5,-1.5) .. (evt.west);
\draw[linkcs]  (udtof.east)   .. controls (3.0,0.3) and (5.5,-2.2) .. (tof.west);

\draw[linkcal] (ourscal.east) .. controls (3.0,-0.8) and (5.5,-2.9) .. (cal.west);

\end{tikzpicture}%
}

\caption{Datasets used for the benchmarking tasks.
$^\dagger$ denotes degradations applied synthetically to clean images from the source dataset.}
\label{fig:dataset-task-grouped}
\end{figure}

The main benchmark spans five categories (Table~\ref{tab:task_summary}). For all subtasks except lens design and calibration, we use a two-stage collection pipeline; data sources are summarized in Fig.~\ref{fig:dataset-task-grouped}.
First, we adopt established datasets when they match the task definition. When an official test split exists, we randomly sample 100 images. Specifically, we use SIDD~\cite{SIDD_2018_CVPR} for denoising, HDR+~\cite{hdrplus_hasinoff2016burst} for HDR reconstruction, MIT-CGH-4K~\cite{shi2021towards} for \CGH, WB Set-1~\cite{afifi2019color} for white balance, Kodak~\cite{li2008image} for demosaicking (Regular/Hexa/Quad), GoPro~\cite{gopro_Nah_2017_CVPR} for motion deblurring, DLMD~\cite{monakhova2019learned} for lensless reconstruction, DAVIS-color~\cite{gao2025unified} for event-based intensity reconstruction, a real under-display \ToF dataset for \ToF imaging, and the Stanford Light Field Dataset~\cite{stanford_lightfield} for light-field view extrapolation.
Second, when no suitable dataset exists, we synthesize measurements using controlled forward models applied to high-quality clean images. For most such subtasks, we build each 100-image set by evenly sampling from DIV2K~\cite{div2k_Agustsson_2017_CVPR_Workshops} and Adobe FiveK~\cite{fivek}, then applying the task-specific forward operator.

For lens design, we construct task instances from target specifications and physical constraints~\cite{wang2022differentiable,yang2024curriculum,chang2018hybrid}. Each optical system is defined by the half-diagonal field of view (HFOV), F-number (FNUM), sensor diagonal (DIAG), number of lens elements, candidate materials, and surface parameterization~\cite{yang2024curriculum}. We fix the sensor diagonal to 6.0\,mm, as well as the number of elements and material list, and parameterize each aspheric surface by one conic constant and six aspheric coefficients. The same aspherical-surface search space is used for all compared lens-design methods. We then vary HFOV and FNUM to control difficulty, yielding benchmark groups that span narrow-to-wide fields of view and slow-to-fast apertures. Calibration task data are collected separately.

\subsection{Ablation Study Data Collection}

For the ablation suite, we generate controlled degradations from clean DIV2K images to isolate specific corruption factors. We consider (i) image noise at multiple levels and (ii) spherical-aberration blur at multiple strengths. For spherical aberration, we inject the Zernike mode \(Z_4^0\)~\cite{wu2019phasecam3d} into the pupil function and vary the aberration strength \(c_{\mathrm{sph}}\) at a fixed wavelength \(\lambda=550\,\mathrm{nm}\). We use \(c_{\mathrm{sph}}\in\{0.1\lambda,0.25\lambda,0.5\lambda,1.0\lambda,1.5\lambda\}\) to synthesize five \PSF kernels, which are then applied to clean images. For denoising, we use a Poisson--Gaussian noise model~\cite{wei2020physics} and sweep ten degradation levels under three settings: Shot-only (vary photon count \(p\), fix read noise \(\sigma\)), Read-only (vary \(\sigma\), fix \(p\)), and Mixed (vary both).

\subsection{Statistical Analysis of the Dataset}

\begin{figure}[t]
  \centering
  \includegraphics[width=\linewidth]{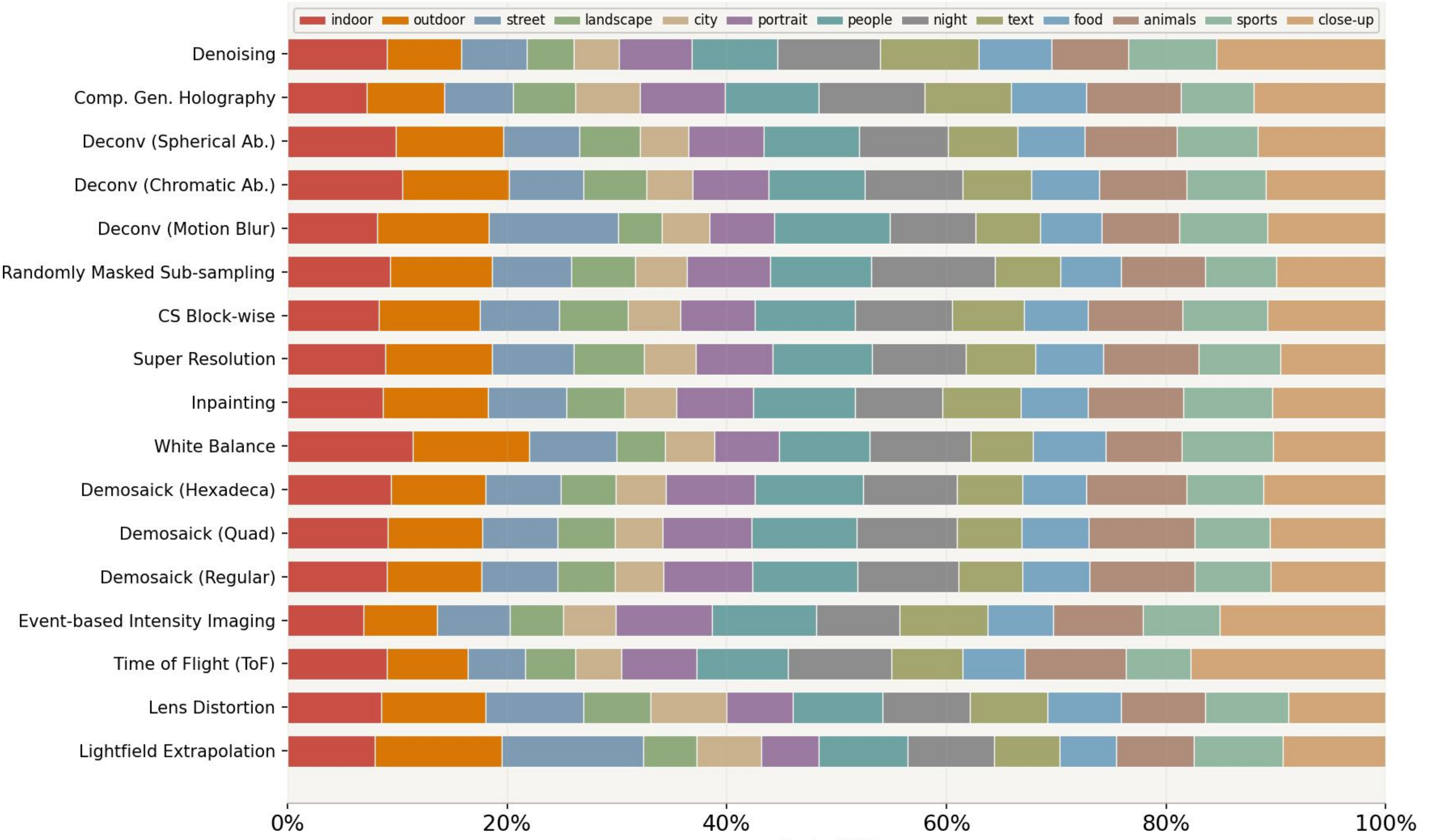}
  \caption{Semantic scene distribution across \ImagingBench subtasks, shown as the average probability over 13 scene categories. The distribution is broadly balanced, with no single scene type dominating the benchmark.}
  \label{fig:distribution}
\end{figure}

\ImagingBench contains 20 subtasks in total (Table~\ref{tab:task_summary}). To characterize dataset diversity, we analyze both semantic scene composition and low-level image statistics. We estimate scene distributions over 13 semantic categories---indoor, outdoor, street, landscape, city, portrait, people, night, text, food, animals, sports, and close-up---and measure brightness and contrast using mean pixel intensity and pixel-intensity standard deviation.
As shown in Fig.~\ref{fig:distribution}, the semantic composition is broadly balanced across subtasks, with no single scene type dominating. The brightness--contrast analysis in the supplementary material likewise shows substantial variation across tasks and samples. Together, these results indicate that \ImagingBench covers diverse semantic content and photometric characteristics, supporting evaluation under varied imaging scenarios.

\section{Results}\label{sec:results}

\begin{figure*}[t]
  \centering
  \includegraphics[width=1.0\linewidth]{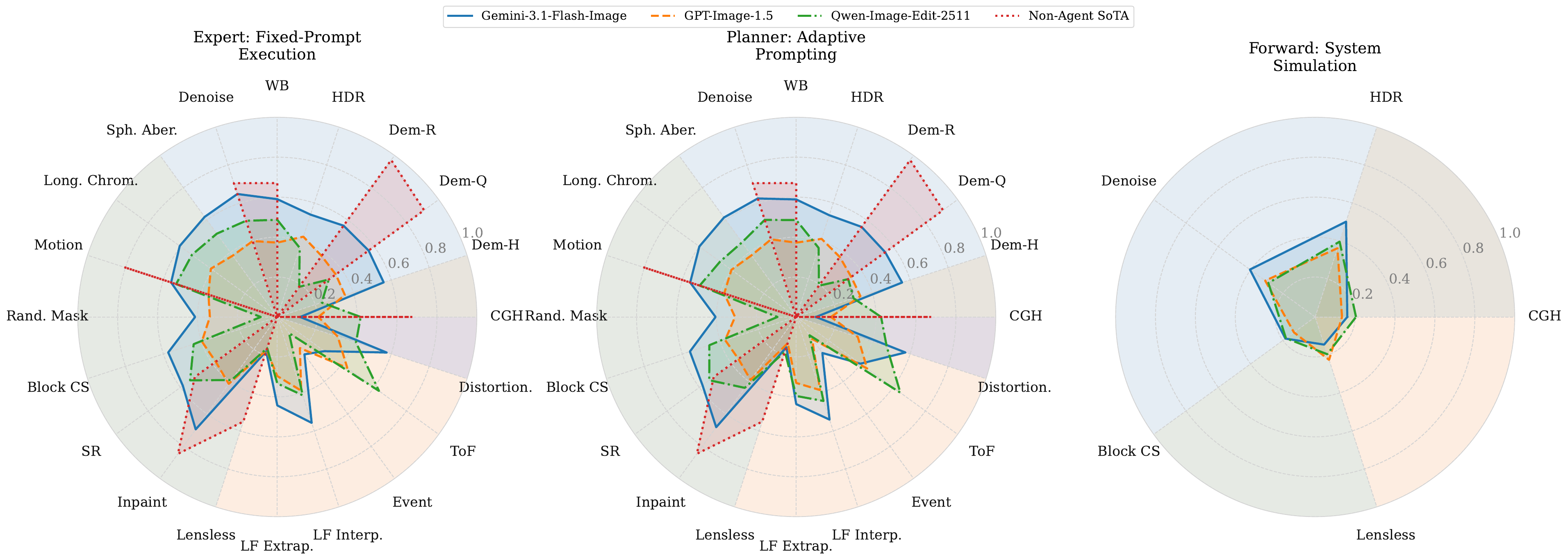}
  \caption{Unified-score radar comparison across representative image-reconstruction subtasks. Each axis reports task-level unified score in $[0,1]$ (higher is better), computed from normalized PSNR/SSIM/LPIPS/NIQE and aggregated with per-instance weighted averaging followed by macro-averaging at task level.}
  \label{fig:unified_scoring_radar}
\end{figure*}

\begin{figure}[t]
  \centering
  \includegraphics[width=0.98\linewidth]{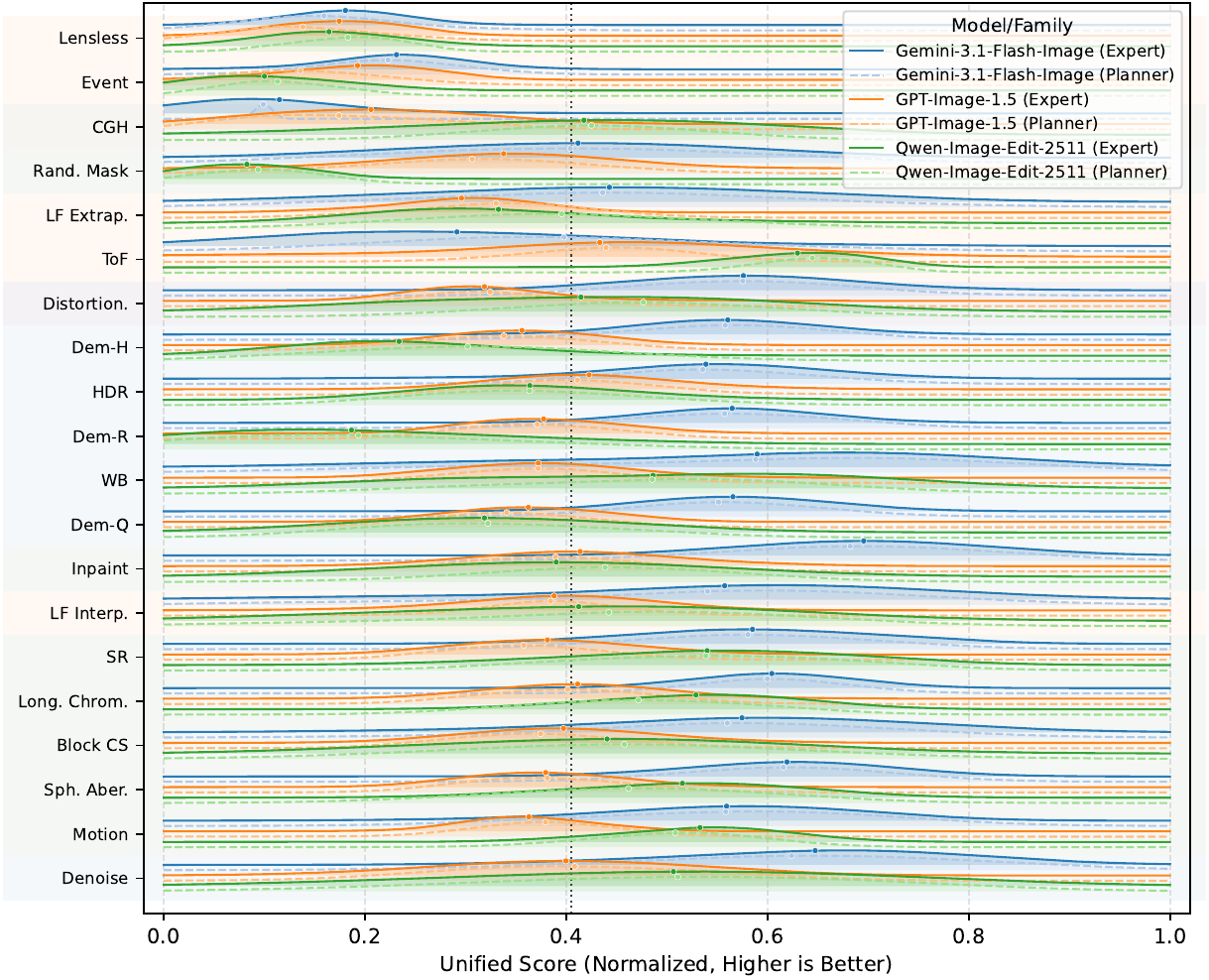}
  \caption{Per-task distributions of the unified score in $[0,1]$ (higher is better) for each model under the Expert and Planner protocols, shown as ridgeline density plots. The unified score is computed from normalized PSNR/SSIM/LPIPS/NIQE with per-instance weighted averaging. Colored dots denote the mean of each distribution.}
  \label{fig:metric_ridge}
\end{figure}

\begin{table*}
\caption{Quantitative results across benchmark tasks for Expert, Planner, and, where available, Forward. In each agent metric cell, the Expert result is shown on the first line, the Planner result is shown on the second line, and the Forward result is shown on the third line only for tasks with Forward evaluation results. Missing Non-agent SoTA phases are shown as \na. Cell colors follow the metric-specific normalization, with green denoting generally considered good and red denoting bad performance for each metric; for stacked cells, each line is colored independently from its own value. }
\label{tab:results_table_m1_m2_stacked}
\centering
\scriptsize
\setlength{\tabcolsep}{3pt}
\renewcommand{\arraystretch}{1.45}
\newcommand{\phasepair}{\vtop{\offinterlineskip\halign{\hfil##\hfil\cr Expert\cr Planner\cr}}}
\newcommand{\phaserows}{\vtop{\offinterlineskip\halign{\hfil##\hfil\cr Expert\cr Planner\cr Forward\cr}}}

\resizebox{\textwidth}{!}{%
\begin{tabular}{l>{\centering\arraybackslash}p{1.3cm}
cccc!{\color{gray!55}\vrule width 0.5pt}
cccc!{\color{gray!55}\vrule width 0.5pt}
cccc!{\color{gray!55}\vrule width 0.5pt}
cccc>{\centering\arraybackslash}p{0.55cm}}
\toprule
\textbf{Task} & \textbf{Phase} &
\multicolumn{4}{c}{\textbf{Gemini-3.1-Flash-Image}} &
\multicolumn{4}{c}{\textbf{GPT-Image-1.5}} &
\multicolumn{4}{c}{\textbf{Qwen-Image-Edit-2511}} &
\multicolumn{5}{c}{\textbf{Non-agent SoTA}} \\

\cmidrule(lr){3-6}
\cmidrule(lr){7-10}
\cmidrule(lr){11-14}
\cmidrule(lr){15-19}

& & PSNR$\uparrow$ & SSIM$\uparrow$ & LPIPS$\downarrow$ & NIQE$\downarrow$
& PSNR$\uparrow$ & SSIM$\uparrow$ & LPIPS$\downarrow$ & NIQE$\downarrow$
& PSNR$\uparrow$ & SSIM$\uparrow$ & LPIPS$\downarrow$ & NIQE$\downarrow$
& PSNR$\uparrow$ & SSIM$\uparrow$ & LPIPS$\downarrow$ & NIQE$\downarrow$ & Cite \\
\midrule

\multicolumn{19}{l}{\cellcolor[gray]{0.9}\textbf{Ray and Wave Optics}} \\
Comp. Gen. Holography
	& \phaserows & \tripsnrcell{6.7390}{5.6094}{10.3545} & \trissimcell{0.0601}{0.0055}{0.1056} & \trilpipscell{1.1179}{1.1238}{0.7632} & \triniqecell{5.6800}{3.4789}{10.8404} & \tripsnrcell{9.5890}{8.1892}{9.9896} & \trissimcell{0.2835}{0.1891}{0.1329} & \trilpipscell{0.8464}{0.9100}{0.9078} & \triniqecell{7.7725}{5.3164}{9.9368} & \tripsnrcell{17.3080}{17.3300}{9.2178} & \trissimcell{0.3867}{0.3354}{0.1597} & \trilpipscell{0.3620}{0.2749}{0.7518} & \triniqecell{6.4624}{6.5748}{6.0035} & \tripsnrcell{25.6180}{\na}{\na} & \trissimcell{0.7715}{\na}{\na} & \trilpipscell{0.2484}{\na}{\na} & \triniqecell{4.3233}{\na}{\na} & \cite{shi2021towards}  \\
\hline
\multicolumn{19}{l}{\cellcolor[gray]{0.9}\textbf{Image Signal Processing}} \\
Demosaicking (Hexadeca)
	& \phasepair & \stackpsnrcell{18.8430}{18.8763} & \stackssimcell{0.6675}{0.6682} & \stacklpipscell{0.2797}{0.2904} & \stackniqecell{3.5547}{3.5765} & \stackpsnrcell{13.6770}{13.8072} & \stackssimcell{0.3635}{0.3405} & \stacklpipscell{0.4995}{0.5365} & \stackniqecell{4.2521}{4.5120} & \stackpsnrcell{12.1980}{14.5206} & \stackssimcell{0.2721}{0.3635} & \stacklpipscell{0.7535}{0.6469} & \stackniqecell{8.0848}{7.1976} & \stackpsnrcell{\na}{\na} & \stackssimcell{\na}{\na} & \stacklpipscell{\na}{\na} & \stackniqecell{\na}{\na} & \na  \\
Demosaicking (Quad)
	& \phasepair & \stackpsnrcell{18.9280}{18.7231} & \stackssimcell{0.6782}{0.6705} & \stacklpipscell{0.2816}{0.2990} & \stackniqecell{3.4191}{3.4598} & \stackpsnrcell{13.8380}{13.9719} & \stackssimcell{0.3629}{0.3375} & \stacklpipscell{0.4836}{0.5249} & \stackniqecell{4.0479}{4.6482} & \stackpsnrcell{14.7440}{15.0206} & \stackssimcell{0.4103}{0.3962} & \stacklpipscell{0.6457}{0.6163} & \stackniqecell{7.2403}{6.9022} & \stackpsnrcell{34.7000}{\na} & \stackssimcell{0.9514}{\na} & \stacklpipscell{0.0397}{\na} & \stackniqecell{3.1143}{\na} & \cite{a2021beyond}  \\
Demosaicking (Regular)
	& \phasepair & \stackpsnrcell{18.8510}{18.8389} & \stackssimcell{0.6844}{0.6770} & \stacklpipscell{0.2831}{0.2898} & \stackniqecell{3.5252}{3.5616} & \stackpsnrcell{14.1850}{14.1506} & \stackssimcell{0.3763}{0.3707} & \stacklpipscell{0.4499}{0.4623} & \stackniqecell{4.1178}{4.1665} & \stackpsnrcell{13.2990}{13.4760} & \stackssimcell{0.2627}{0.2699} & \stacklpipscell{0.7905}{0.7820} & \stackniqecell{14.9870}{14.8543} & \stackpsnrcell{38.7200}{\na} & \stackssimcell{0.9707}{\na} & \stacklpipscell{0.0165}{\na} & \stackniqecell{2.9548}{\na} & \cite{a2021beyond}  \\
High Dynamic Range
	& \phaserows & \tripsnrcell{18.5050}{17.7578}{17.9266} & \trissimcell{0.5844}{0.5751}{0.5416} & \trilpipscell{0.2092}{0.2145}{0.3158} & \triniqecell{3.7770}{3.7768}{4.1086} & \tripsnrcell{15.7260}{15.2373}{13.8014} & \trissimcell{0.4406}{0.4302}{0.3934} & \trilpipscell{0.3944}{0.4071}{0.5180} & \triniqecell{4.1285}{4.2278}{4.0377} & \tripsnrcell{13.4110}{13.4061}{13.3759} & \trissimcell{0.3878}{0.3978}{0.4739} & \trilpipscell{0.4782}{0.4831}{0.4712} & \triniqecell{5.6890}{6.0185}{4.6475} & \tripsnrcell{\na}{\na}{\na} & \trissimcell{\na}{\na}{\na} & \trilpipscell{\na}{\na}{\na} & \triniqecell{\na}{\na}{\na} & \na  \\
White Balance
& \phasepair & \stackpsnrcell{18.9970}{19.3164} & \stackssimcell{0.7510}{0.7460} & \stacklpipscell{0.3108}{0.3191} & \stackniqecell{4.2118}{4.3346} & \stackpsnrcell{11.9610}{11.8443} & \stackssimcell{0.4352}{0.4392} & \stacklpipscell{0.5120}{0.5169} & \stackniqecell{3.9845}{4.0529} & \stackpsnrcell{15.2340}{15.4550} & \stackssimcell{0.6589}{0.6627} & \stacklpipscell{0.4298}{0.4234} & \stackniqecell{5.0550}{5.0137} & \stackpsnrcell{25.1660}{\na} & \stackssimcell{0.7983}{\na} & \stacklpipscell{0.2753}{\na} & \stackniqecell{4.4722}{\na} & \cite{afifi2020deepWB}  \\
Denoising
	& \phaserows & \tripsnrcell{25.4330}{23.8154}{19.5774} & \trissimcell{0.7950}{0.7779}{0.4410} & \trilpipscell{0.3118}{0.3143}{0.5693} & \triniqecell{6.9354}{6.9780}{6.0562} & \tripsnrcell{18.0020}{17.9841}{17.7168} & \trissimcell{0.5176}{0.5424}{0.2556} & \trilpipscell{0.5526}{0.5391}{0.5899} & \triniqecell{8.9955}{9.5650}{7.4043} & \tripsnrcell{20.1890}{20.9247}{17.9078} & \trissimcell{0.6880}{0.6689}{0.4167} & \trilpipscell{0.4705}{0.4428}{0.8889} & \triniqecell{8.1577}{8.3589}{8.7536} & \tripsnrcell{32.8350}{\na}{\na} & \trissimcell{0.7542}{\na}{\na} & \trilpipscell{0.4126}{\na}{\na} & \triniqecell{5.0361}{\na}{\na} & \cite{zhang2018ffdnet}  \\
\hline
\multicolumn{19}{l}{\cellcolor[gray]{0.9}\textbf{Inverse Reconstruction}} \\
Spherical Aberration
	& \phasepair & \stackpsnrcell{22.6920}{22.6349} & \stackssimcell{0.6454}{0.6459} & \stacklpipscell{0.2049}{0.2106} & \stackniqecell{3.8153}{3.8884} & \stackpsnrcell{14.2160}{14.4354} & \stackssimcell{0.3854}{0.3900} & \stacklpipscell{0.4564}{0.4576} & \stackniqecell{4.0016}{4.2209} & \stackpsnrcell{20.6100}{19.5474} & \stackssimcell{0.6652}{0.6055} & \stacklpipscell{0.2965}{0.4791} & \stackniqecell{5.1592}{8.3401} & \stackpsnrcell{\na}{\na} & \stackssimcell{\na}{\na} & \stacklpipscell{\na}{\na} & \stackniqecell{\na}{\na} & \na  \\
Longitudinal Chromatic
	& \phasepair & \stackpsnrcell{21.5190}{21.4794} & \stackssimcell{0.6720}{0.6739} & \stacklpipscell{0.2317}{0.2444} & \stackniqecell{3.9753}{4.0707} & \stackpsnrcell{15.7070}{15.5315} & \stackssimcell{0.4674}{0.4537} & \stacklpipscell{0.4528}{0.4701} & \stackniqecell{4.8180}{4.8220} & \stackpsnrcell{19.6200}{18.9918} & \stackssimcell{0.6612}{0.5934} & \stacklpipscell{0.3745}{0.4585} & \stackniqecell{5.2546}{6.2080} & \stackpsnrcell{\na}{\na} & \stackssimcell{\na}{\na} & \stacklpipscell{\na}{\na} & \stackniqecell{\na}{\na} & \na  \\
Motion Blur
& \phasepair & \stackpsnrcell{21.7250}{21.6717} & \stackssimcell{0.6557}{0.6574} & \stacklpipscell{0.3713}{0.3665} & \stackniqecell{3.6723}{3.9662} & \stackpsnrcell{14.6550}{15.1555} & \stackssimcell{0.4277}{0.4636} & \stacklpipscell{0.5443}{0.5218} & \stackniqecell{4.2063}{4.9140} & \stackpsnrcell{20.6970}{20.4214} & \stackssimcell{0.7620}{0.7361} & \stacklpipscell{0.4245}{0.4572} & \stackniqecell{9.2441}{9.9118} & \stackpsnrcell{31.3730}{\na} & \stackssimcell{0.9174}{\na} & \stacklpipscell{0.1363}{\na} & \stackniqecell{6.9446}{\na} & \cite{Tsai2022BANet}  \\
Randomly Masked Sub-Sampling
	& \phasepair & \stackpsnrcell{13.9600}{13.6010} & \stackssimcell{0.4733}{0.4651} & \stacklpipscell{0.4621}{0.4747} & \stackniqecell{4.4147}{4.8592} & \stackpsnrcell{12.9900}{11.9729} & \stackssimcell{0.3656}{0.3368} & \stacklpipscell{0.5554}{0.6123} & \stackniqecell{4.8011}{5.4764} & \stackpsnrcell{9.4000}{8.8198} & \stackssimcell{0.0888}{0.1154} & \stacklpipscell{0.9742}{0.9315} & \stackniqecell{16.3520}{11.2581} & \stackpsnrcell{\na}{\na} & \stackssimcell{\na}{\na} & \stacklpipscell{\na}{\na} & \stackniqecell{\na}{\na} & \na  \\
Comp. Sensing (Block)
	& \phaserows & \tripsnrcell{22.4500}{21.3091}{10.5505} & \trissimcell{0.6206}{0.6180}{0.2452} & \trilpipscell{0.2544}{0.2635}{0.7941} & \triniqecell{3.7715}{3.7447}{11.7572} & \tripsnrcell{14.9270}{13.7650}{9.9144} & \trissimcell{0.4228}{0.3944}{0.2177} & \trilpipscell{0.4466}{0.4784}{0.8197} & \triniqecell{4.0858}{4.2230}{20.2304} & \tripsnrcell{20.9340}{21.6684}{9.0825} & \trissimcell{0.6274}{0.6396}{0.1820} & \trilpipscell{0.6056}{0.5816}{0.7019} & \triniqecell{10.1280}{10.0400}{17.5821} & \tripsnrcell{\na}{\na}{\na} & \trissimcell{\na}{\na}{\na} & \trilpipscell{\na}{\na}{\na} & \triniqecell{\na}{\na}{\na} & \na  \\
Super-Resolution
	& \phasepair & \stackpsnrcell{21.7450}{21.4739} & \stackssimcell{0.6049}{0.5994} & \stacklpipscell{0.2400}{0.2424} & \stackniqecell{4.0156}{3.8287} & \stackpsnrcell{14.3650}{13.7717} & \stackssimcell{0.3915}{0.3658} & \stacklpipscell{0.4602}{0.5021} & \stackniqecell{4.0148}{4.0204} & \stackpsnrcell{20.1560}{19.8193} & \stackssimcell{0.6122}{0.5990} & \stacklpipscell{0.3229}{0.3047} & \stackniqecell{4.8301}{4.6428} & \stackpsnrcell{20.5480}{\na} & \stackssimcell{0.5811}{\na} & \stacklpipscell{0.3856}{\na} & \stackniqecell{5.4237}{\na} & \cite{wang2018esrgan}  \\
Inpainting
	& \phasepair & \stackpsnrcell{23.7770}{23.1095} & \stackssimcell{0.7761}{0.7667} & \stacklpipscell{0.1242}{0.1365} & \stackniqecell{3.8433}{3.8050} & \stackpsnrcell{15.1800}{14.2718} & \stackssimcell{0.4347}{0.4186} & \stacklpipscell{0.4069}{0.4541} & \stackniqecell{4.2309}{4.1996} & \stackpsnrcell{12.8130}{13.9815} & \stackssimcell{0.4744}{0.5337} & \stacklpipscell{0.4934}{0.4069} & \stackniqecell{5.3394}{4.9925} & \stackpsnrcell{30.3260}{\na} & \stackssimcell{0.9274}{\na} & \stacklpipscell{0.0508}{\na} & \stackniqecell{3.0213}{\na} & \cite{suvorov2022resolution}  \\
\hline
\multicolumn{19}{l}{\cellcolor[gray]{0.9}\textbf{Computational Sensing}} \\
Lensless Camera
	& \phaserows & \tripsnrcell{7.4770}{7.2833}{8.8316} & \trissimcell{0.2112}{0.1887}{0.1530} & \trilpipscell{0.7875}{0.7917}{0.9481} & \triniqecell{11.3900}{16.8795}{5.8861} & \tripsnrcell{7.9040}{8.0157}{10.0590} & \trissimcell{0.2349}{0.2429}{0.3291} & \trilpipscell{0.8373}{0.8875}{0.7898} & \triniqecell{11.0680}{16.2779}{9.1535} & \tripsnrcell{8.9740}{8.8098}{7.9138} & \trissimcell{0.2518}{0.2415}{0.2610} & \trilpipscell{0.7931}{0.7627}{0.8280} & \triniqecell{16.2590}{13.4506}{8.7611} & \tripsnrcell{19.9800}{\na}{\na} & \trissimcell{0.6777}{\na}{\na} & \trilpipscell{0.2885}{\na}{\na} & \triniqecell{7.3787}{\na}{\na} & \cite{monakhova2019learned}  \\
Lightfield View Extrapolation
& \phasepair & \stackpsnrcell{16.4193}{16.3840} & \stackssimcell{0.5315}{0.5217} & \stacklpipscell{0.4511}{0.4544} & \stackniqecell{5.4893}{5.1278} & \stackpsnrcell{12.3015}{14.2579} & \stackssimcell{0.3514}{0.3698} & \stacklpipscell{0.6690}{0.6043} & \stackniqecell{4.6696}{4.3547} & \stackpsnrcell{11.8973}{15.4003} & \stackssimcell{0.4336}{0.5057} & \stacklpipscell{0.6269}{0.5256} & \stackniqecell{6.6586}{6.7052} & \stackpsnrcell{\na}{\na} & \stackssimcell{\na}{\na} & \stacklpipscell{\na}{\na} & \stackniqecell{\na}{\na} & \na  \\
Lightfield View Interpolation
& \phasepair & \stackpsnrcell{21.2975}{20.8270} & \stackssimcell{0.6366}{0.6144} & \stacklpipscell{0.3279}{0.3492} & \stackniqecell{5.8781}{5.6365} & \stackpsnrcell{15.4606}{15.5112} & \stackssimcell{0.4679}{0.4497} & \stacklpipscell{0.5054}{0.5090} & \stackniqecell{5.6446}{4.6344} & \stackpsnrcell{16.5642}{18.3563} & \stackssimcell{0.5338}{0.5531} & \stacklpipscell{0.5294}{0.4825} & \stackniqecell{8.7182}{8.6902} & \stackpsnrcell{\na}{\na} & \stackssimcell{\na}{\na} & \stacklpipscell{\na}{\na} & \stackniqecell{\na}{\na} & \na  \\
Event-based Intensity Imaging
& \phasepair & \stackpsnrcell{9.9510}{10.0585} & \stackssimcell{0.2943}{0.2677} & \stacklpipscell{0.7345}{0.7392} & \stackniqecell{9.4659}{9.3335} & \stackpsnrcell{10.0300}{10.2295} & \stackssimcell{0.2305}{0.1491} & \stacklpipscell{0.7978}{0.8896} & \stackniqecell{9.5875}{11.5704} & \stackpsnrcell{7.1170}{7.5260} & \stackssimcell{0.1124}{0.0682} & \stacklpipscell{0.8602}{0.7741} & \stackniqecell{28.3040}{22.2731} & \stackpsnrcell{\na}{\na} & \stackssimcell{\na}{\na} & \stacklpipscell{\na}{\na} & \stackniqecell{\na}{\na} & \na  \\
Time-of-Flight Depth Imaging
& \phasepair & \stackpsnrcell{4.9900}{10.4987} & \stackssimcell{0.0704}{0.5251} & \stacklpipscell{0.6796}{0.4341} & \stackniqecell{9.1302}{9.7917} & \stackpsnrcell{9.4150}{12.6478} & \stackssimcell{0.3518}{0.6369} & \stacklpipscell{0.5519}{0.3579} & \stackniqecell{11.7790}{11.2442} & \stackpsnrcell{23.9250}{23.8948} & \stackssimcell{0.7956}{0.7809} & \stacklpipscell{0.0780}{0.0754} & \stackniqecell{15.6590}{15.4957} & \stackpsnrcell{\na}{\na} & \stackssimcell{\na}{\na} & \stacklpipscell{\na}{\na} & \stackniqecell{\na}{\na} & \na  \\
\hline
\multicolumn{19}{l}{\cellcolor[gray]{0.9}\textbf{Calibration}} \\
Camera Distortion Score
& \phasepair & \stackpsnrcell{19.1297}{19.0303} & \stackssimcell{0.5606}{0.5622} & \stacklpipscell{0.1317}{0.1305} & \stackniqecell{3.2347}{3.2595} & \stackpsnrcell{11.0319}{11.2836} & \stackssimcell{0.2971}{0.3044} & \stacklpipscell{0.5526}{0.5385} & \stackniqecell{3.8971}{3.9590} & \stackpsnrcell{14.5423}{17.0120} & \stackssimcell{0.4332}{0.4776} & \stacklpipscell{0.4194}{0.2948} & \stackniqecell{4.4507}{4.2799} & \stackpsnrcell{\na}{\na} & \stackssimcell{\na}{\na} & \stacklpipscell{\na}{\na} & \stackniqecell{\na}{\na} & \na  \\

\bottomrule
\end{tabular}%
}
\end{table*}

\begin{table}[t]
\centering
\caption{\ImagingBench{} leaderboard under the Expert protocol. \textbf{Overall} is the category-weighted score (macro-average of the five task categories); \textbf{Agg} is the aggregate score under a flat per-task ($1/|\mathcal{T}|$) weighting. Reconstruction tasks contribute through their categories; camera calibration enters the Calibration category and lens design the Ray-and-Wave-Optics category. Higher is better; the per-category breakdown is in the supplement.}
\label{tab:overall_scores}
\begin{tabular}{lcc}
\toprule
\textbf{Model} & \textbf{Overall} & \textbf{Agg} \\
\midrule
Gemini-3.1-Flash-Image & 0.413 & 0.471 \\
GPT-Image-1.5 & 0.305 & 0.336 \\
Qwen-Image-Edit-2511 & 0.341 & 0.362 \\
\bottomrule
\end{tabular}
\end{table}

\subsection{Overall Quantitative Trends}

Table~\ref{tab:results_table_m1_m2_stacked} reports raw task-level results for Expert and Planner, Fig.~\ref{fig:metric_ridge} summarizes the corresponding metric distributions, and Fig.~\ref{fig:unified_scoring_radar} shows task-level unified scores. Together, they reveal several consistent trends.

\noindent \textbf{Agentic vs. non-agentic.}
Where specialized non-agentic baselines are available, they generally outperform the agentic multimodal systems. This is clear in Table~\ref{tab:results_table_m1_m2_stacked}, where the baseline columns are often stronger, and in the radar plots, where the non-agentic curve frequently lies outside the agentic ones. Thus, general agentic multimodal capability still falls short of specialized non-agentic methods.

\noindent \textbf{Fidelity vs. plausibility.}
The metrics show a clear mismatch between visual plausibility and reconstruction fidelity. PSNR is often low for agentic models, and LPIPS is also generally worse than the specialized baselines, indicating poor agreement with the target. SSIM is usually less degraded, suggesting that coarse scene structure is often preserved. By contrast, NIQE is sometimes relatively favorable even when PSNR and LPIPS are poor, implying that the outputs may look natural without being correct.

\noindent \textbf{Task-family differences.}
Agentic models are more competitive on conventional image signal processing and inverse reconstruction tasks, such as denoising, white balance, inpainting, super-resolution, and some blur-removal problems. They perform much worse on computational sensing tasks, including lensless imaging, event-based intensity imaging, time-of-flight depth imaging, and holography. This trend appears in both the raw table and the unified-score radar plots, indicating that structured sensing operators remain substantially harder than photographic degradations.

\noindent \textbf{Expert vs. Planner.}
Planner guidance provides only modest and inconsistent gains. In our setup, Expert uses a fixed expert prompt, whereas Planner instantiates the observe--plan--execute pipeline by having a planner generate an image-specific restoration instruction before execution. In some tasks, Planner slightly improves SSIM, LPIPS, or NIQE over Expert, suggesting that image-specific prompting can help when the degradation is correctly diagnosed. However, these gains are usually small and do not change the overall ranking. The average unified-score delta from Expert to Planner is also small across model families: $-0.0009$ for Gemini, $-0.0103$ for GPT, and $+0.0109$ for Qwen. These near-zero means suggest that prompt sensitivity is modest on average rather than dominant, although task-wise gains and losses can cancel. The radar plots for Expert and Planner therefore retain similar shapes, showing that planner guidance helps only in selected cases.

\noindent \textbf{Model-family differences.}
Across both Expert and Planner, Gemini is generally the strongest and most consistent family on standard restoration tasks, GPT is usually weaker, and Qwen is the most variable. Qwen occasionally becomes competitive on sensing-oriented tasks, especially time-of-flight depth imaging, but is less stable overall. This suggests that base-model capability matters more than planner design alone.

\noindent \textbf{Overall score.}
Table~\ref{tab:overall_scores} is the \ImagingBench{} leaderboard under the Expert
protocol. It reports two aggregate scores per model: the category-weighted
\emph{Overall} (macro-average of the five task-category scores) and the
\emph{Agg} score under a flat per-task ($1/|\mathcal{T}|$) weighting.
Reconstruction tasks contribute through their categories, while camera
calibration (scored by intrinsic/distortion parameter error) and lens design
(scored by normalized Strehl ratio) enter the Calibration and optics categories,
so both scores reflect non-reconstruction competence as well. The ranking mirrors
the per-task trends and is identical under either weighting: Gemini leads
($0.41$ / $0.47$), followed by Qwen ($0.34$ / $0.36$) and GPT ($0.31$ / $0.34$),
confirming that stronger base multimodal capability, rather than protocol choice
or aggregation scheme, drives benchmark standing. A per-category breakdown is
given in the supplement.

\noindent \textbf{Summary.}
Overall, the results support three conclusions: specialized non-agentic methods remain substantially stronger than current agentic multimodal systems; computational sensing tasks are markedly harder than conventional restoration tasks; and visually plausible outputs do not necessarily correspond to physically accurate solutions.

\subsection{Individual Task Results}

Due to space constraints, we show representative individual task results here and provide additional results in the supplementary material.

\begin{figure}[t]
  \centering
  \includegraphics[width=1\linewidth]{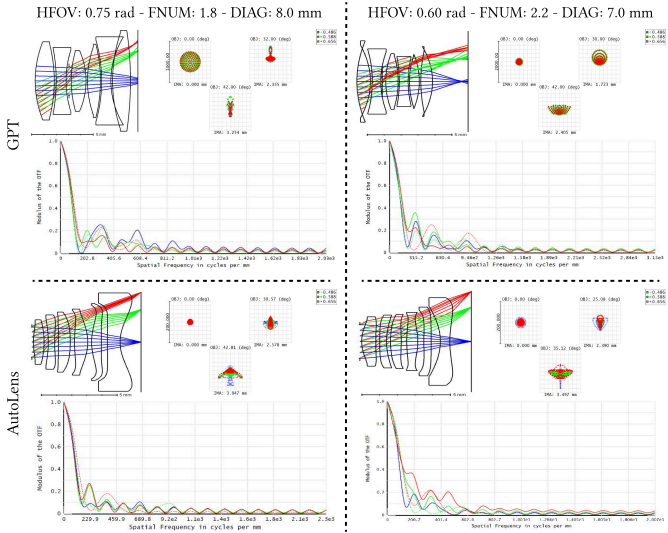}
  \caption{Comparison of representative ray-optics lens designs under different target specifications. From left to right, the three columns correspond to benchmark groups with decreasing design difficulty, moving from wider fields of view, faster apertures, and larger sensor formats to narrower fields of view, slower apertures, and smaller sensor formats. For each design, we show the optical layout, spot diagrams at multiple field angles, and the corresponding MTF curves. The top and bottom rows present the results of \textbf{GPT-5.4} and \textbf{AutoLens}~\cite{yang2024curriculum}, respectively, highlighting the gap between a general \VLM and a state-of-the-art learning-based lens design method in both structural plausibility and optical performance.}
  \label{fig:lens_design_auto_vs_gpt}
\end{figure}

\noindent \textbf{Lens Design} Fig.~\ref{fig:lens_design_auto_vs_gpt} compares ChatGPT with AutoLens~\cite{yang2024curriculum} on three representative lens design targets. The corresponding results for Gemini and Qwen are provided in the supplementary material. Both methods are evaluated under the same aspherical-surface parameterization and search space. From both the spot diagrams and the MTF curves, AutoLens outperforms ChatGPT on most targets. Specifically, AutoLens generally produces tighter and better-corrected spot patterns with less chromatic spread, while ChatGPT more often generates broader spot distributions, stronger color separation, and weaker off-axis aberration correction. The MTF curves further confirm that AutoLens preserves higher and more stable modulation transfer. Although ChatGPT can generate optically valid lens layouts that satisfy the basic system specifications, its final optical performance still requires further improvement. Accordingly, this task should be interpreted as testing whether a general \VLM can propose a useful one-shot initial optical design, not whether it can replace a full iterative optical-design pipeline. Overall, these results indicate that current LLM-based lens design is better suited to generating coarse initial solutions, which may serve as useful starting points for subsequent refinement, rather than directly producing high-quality final optical designs. Quantitatively, the peak on-axis Strehl ratio (where $1.0$ denotes the ideal diffraction-limited value) averaged over the five targets is $0.023$ for AutoLens---the highest among all methods and the leader on four of the five targets---versus $0.014$ for GPT-5.4 and $0.013$ for Gemini, while Qwen produces an analyzable prescription for only a single target ($0.005$). All designs remain far below the $0.8$ diffraction-limited threshold, reinforcing that current \VLM lens prescriptions are at best coarse initializations; per-target Strehl values are reported in the supplementary material.

\begin{figure*}[t]
  \centering
  \includegraphics[width=1.0\linewidth]{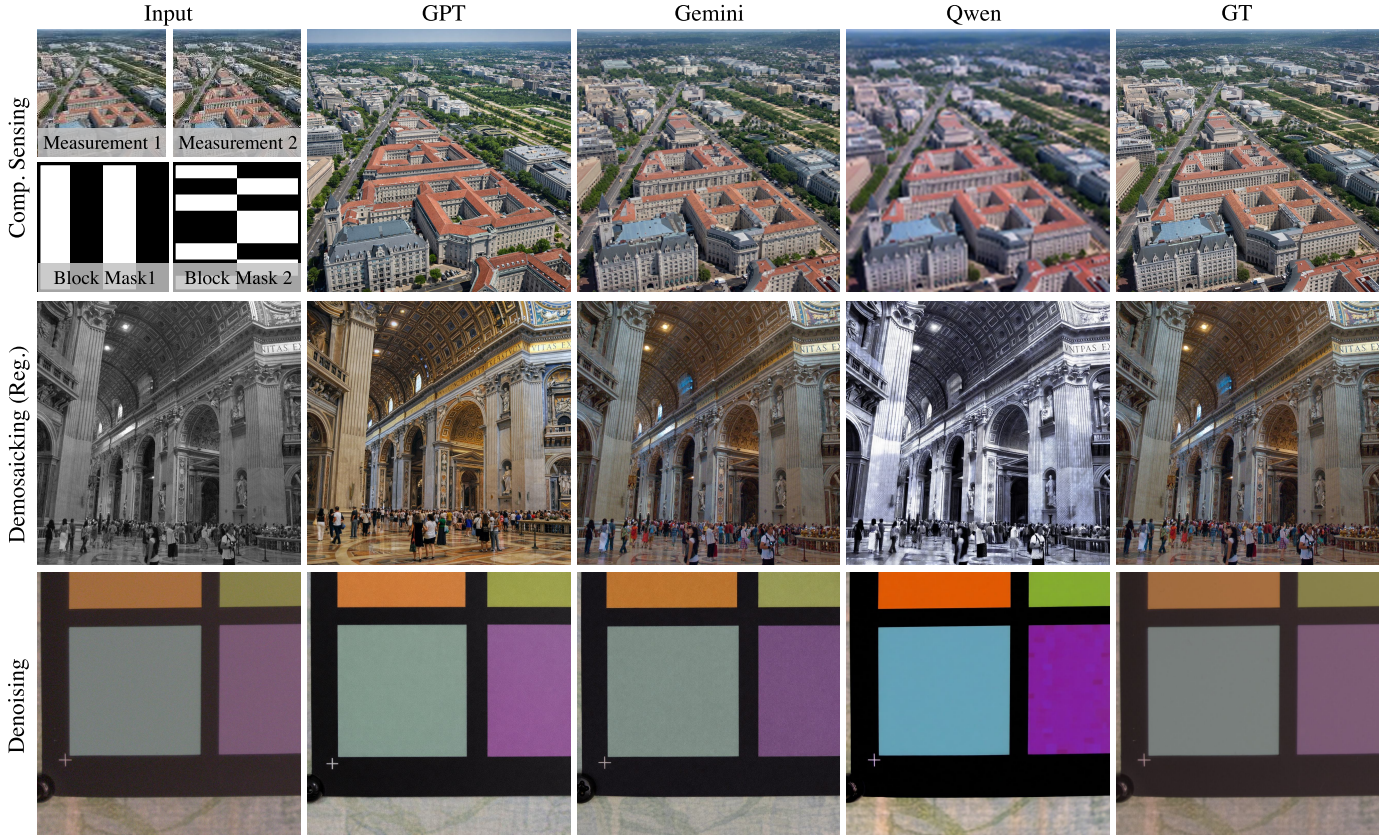}
  \caption{Visual comparison of outputs from GPT, Gemini, and Qwen on three representative ImagingBench tasks: compressive sensing (top), demosaicking (middle), and denoising (bottom). The examples highlight differences in reconstruction fidelity across models in terms of structural recovery, color restoration, and noise suppression. Additional visual results are provided in the supplementary material.}
  \label{fig:main_visual_results}
\end{figure*}

\noindent \textbf{Representative visual reconstruction results.} Fig.~\ref{fig:main_visual_results} shows three representative image tasks: compressive sensing, demosaicking, and denoising. The examples mirror the overall quantitative trends. On demosaicking and denoising, the models usually recover the global scene structure, but still exhibit color errors, texture loss, or oversmoothing relative to the ground truth. On compressive sensing, missing information is often filled in with visually plausible but incorrect content, indicating weak operator-aware inversion. Across the three examples, Gemini is generally the most faithful, GPT is weaker, and Qwen is more variable, consistent with the broader benchmark statistics.

\section{Discussion and Analysis}

\subsection{Sampling-Ratio Ablation}

Table~\ref{tab:cs_ratio} studies how performance changes as the sampling ratio varies. As expected, performance improves as more information is retained, but the gain is not linear and remains model dependent. At aggressive undersampling, the models often produce visually plausible outputs that are inconsistent with the target, indicating that current agentic systems still require substantially more information than specialized inverse methods to achieve reliable recovery.

\begin{table}[t]
\caption{Reconstruction performance under pixel-space compressive sensing with different sampling ratios (fraction of retained pixels).}
\label{tab:cs_ratio}
\centering
\scriptsize
\setlength{\tabcolsep}{3pt}
\renewcommand{\arraystretch}{1.08}
\begin{tabular}{llccccc}
\toprule
Model & Metric & 10\% & 20\% & 40\% & 60\% & 80\% \\
\midrule

\multirow{4}{*}{Gemini}
& PSNR$\uparrow$    & 8.9235 & 11.066 & 13.459 & 16.652 & 19.650 \\
& SSIM$\uparrow$    & 0.2909 & 0.3924 & 0.4878 & 0.5707 & 0.6228 \\
& LPIPS$\downarrow$ & 0.7158 & 0.5496 & 0.4328 & 0.3347 & 0.2800 \\
& NIQE$\downarrow$  & 6.2349 & 4.7583 & 3.9246 & 3.6435 & 3.5308 \\
\midrule

\multirow{4}{*}{GPT}
& PSNR$\uparrow$    & 10.450 & 11.374 & 13.764 & 14.545 & 14.936 \\
& SSIM$\uparrow$    & 0.3180 & 0.3312 & 0.3827 & 0.3928 & 0.4053 \\
& LPIPS$\downarrow$ & 0.6960 & 0.6144 & 0.5250 & 0.4804 & 0.4549 \\
& NIQE$\downarrow$  & 5.2085 & 5.0434 & 4.6828 & 4.4991 & 4.5530 \\
\midrule

\multirow{4}{*}{Qwen}
& PSNR$\uparrow$    & 6.4209 & 8.0802 & 11.105 & 9.9738 & 11.422 \\
& SSIM$\uparrow$    & 0.0432 & 0.1037 & 0.1318 & 0.0770 & 0.0885 \\
& LPIPS$\downarrow$ & 0.9721 & 0.9320 & 1.0096 & 1.0514 & 0.9060 \\
& NIQE$\downarrow$  & 18.931 & 10.072 & 10.192 & 18.007 & 24.556 \\
\bottomrule
\end{tabular}
\end{table}

\subsection{Noise Ablation}

For sensing tasks, noise tolerance is a basic empirical question because reconstruction quality depends on how much signal survives the measurement process. We therefore perform an ablation study across ten degradation levels under three noise settings, as detailed in Table~\ref{tab:noise_3cases}. 
\textbf{Case 1 (Shot Noise)} fixes $\sigma=0.2157$ while decreasing $p$ from 1000 to 5 to isolate signal-dependent uncertainty; \textbf{Case 2 (Read Noise)} fixes $p=70.7$ while increasing $\sigma$ from 0.0392 to 0.3922 to test resilience against constant electronic interference; and \textbf{Case 3 (Mixed Noise)} varies both $p$ and $\sigma$ simultaneously to simulate realistic low-light conditions. Gemini is substantially more robust than GPT and Qwen under moderate to extreme noise.

\begingroup
\renewcommand{\psnrcell}[1]{\heatup{#1}{12}{27}}
\renewcommand{\ssimcell}[1]{\heatup{#1}{0.30}{0.80}}
\renewcommand{\lpipscell}[1]{\heatdown{#1}{0.12}{0.70}}
\renewcommand{\niqecell}[1]{\heatdown{#1}{3.5}{6.1}}
\sisetup{round-mode=places,round-precision=3,group-digits=false}
\begin{table}[t]
\caption{Denoising performance under three noise-setting cases. Cell colors follow the same higher-is-better / lower-is-better convention as Table~\ref{tab:results_table_m1_m2_stacked}, but are normalized to the range of this ablation study.}
\label{tab:noise_3cases}
\centering
\scriptsize
\setlength{\tabcolsep}{2pt}
\renewcommand{\arraystretch}{1.15}
\resizebox{\columnwidth}{!}{%
\begin{tabular}{c c c
                !{\color{gray!55}\vrule width 0.5pt}
                cccc
                !{\color{gray!55}\vrule width 0.5pt}
                cccc
                !{\color{gray!55}\vrule width 0.5pt}
                cccc}
\toprule
Level & {$p$} & {$\sigma$} &
\multicolumn{4}{c}{\makecell{Gemini-3.1\\Flash-Image}} &
\multicolumn{4}{c}{GPT-Image-1.5} &
\multicolumn{4}{c}{\makecell{Qwen-Image\\Edit-2511}} \\
\cmidrule(lr){4-7}\cmidrule(lr){8-11}\cmidrule(lr){12-15}
 & & &
\makecell{PSNR\\$\uparrow$} & \makecell{SSIM\\$\uparrow$} & \makecell{LPIPS\\$\downarrow$} & \makecell{NIQE\\$\downarrow$} &
\makecell{PSNR\\$\uparrow$} & \makecell{SSIM\\$\uparrow$} & \makecell{LPIPS\\$\downarrow$} & \makecell{NIQE\\$\downarrow$} &
\makecell{PSNR\\$\uparrow$} & \makecell{SSIM\\$\uparrow$} & \makecell{LPIPS\\$\downarrow$} & \makecell{NIQE\\$\downarrow$} \\
\midrule

\multicolumn{15}{l}{\cellcolor[gray]{0.92}\textbf{Case 1 (Shot varies, Read fixed):} $\sigma=\num{0.2157}$ fixed, $p$ varies} \\
0 & 1000.0 & 0.2157 & \psnrcell{22.454} & \ssimcell{0.6345} & \lpipscell{0.2647} & \niqecell{3.577} & \psnrcell{14.951} & \ssimcell{0.3439} & \lpipscell{0.4659} & \niqecell{4.491} & \psnrcell{18.516} & \ssimcell{0.6045} & \lpipscell{0.3930} & \niqecell{4.744} \\
2 & 308.0 & 0.2157 & \psnrcell{21.820} & \ssimcell{0.6242} & \lpipscell{0.2730} & \niqecell{3.827} & \psnrcell{15.250} & \ssimcell{0.3440} & \lpipscell{0.4599} & \niqecell{4.768} & \psnrcell{19.152} & \ssimcell{0.6086} & \lpipscell{0.3892} & \niqecell{5.051} \\
4 & 94.9 & 0.2157 & \psnrcell{22.093} & \ssimcell{0.6142} & \lpipscell{0.2659} & \niqecell{3.763} & \psnrcell{14.884} & \ssimcell{0.3549} & \lpipscell{0.4681} & \niqecell{4.769} & \psnrcell{18.627} & \ssimcell{0.6162} & \lpipscell{0.3724} & \niqecell{4.901} \\
6 & 29.2 & 0.2157 & \psnrcell{21.629} & \ssimcell{0.6048} & \lpipscell{0.2774} & \niqecell{3.768} & \psnrcell{14.905} & \ssimcell{0.3389} & \lpipscell{0.4604} & \niqecell{4.610} & \psnrcell{18.104} & \ssimcell{0.5533} & \lpipscell{0.4715} & \niqecell{4.947} \\
8 & 9.0 & 0.2157 & \psnrcell{20.047} & \ssimcell{0.5747} & \lpipscell{0.3033} & \niqecell{3.517} & \psnrcell{14.863} & \ssimcell{0.3377} & \lpipscell{0.4716} & \niqecell{4.548} & \psnrcell{15.853} & \ssimcell{0.5007} & \lpipscell{0.5424} & \niqecell{4.996} \\
\midrule

\multicolumn{15}{l}{\cellcolor[gray]{0.92}\textbf{Case 2 (Read varies, Shot fixed):} $p=\num{70.7}$ fixed, $\sigma$ varies} \\
0 & 70.7 & 0.0392 & \psnrcell{25.044} & \ssimcell{0.7431} & \lpipscell{0.1650} & \niqecell{3.753} & \psnrcell{15.355} & \ssimcell{0.3543} & \lpipscell{0.4598} & \niqecell{4.629} & \psnrcell{21.131} & \ssimcell{0.7413} & \lpipscell{0.2108} & \niqecell{4.168} \\
2 & 70.7 & 0.1176 & \psnrcell{23.905} & \ssimcell{0.6745} & \lpipscell{0.2087} & \niqecell{3.559} & \psnrcell{15.491} & \ssimcell{0.3613} & \lpipscell{0.4643} & \niqecell{4.746} & \psnrcell{20.378} & \ssimcell{0.6977} & \lpipscell{0.2637} & \niqecell{4.755} \\
4 & 70.7 & 0.1961 & \psnrcell{22.459} & \ssimcell{0.6368} & \lpipscell{0.2589} & \niqecell{3.629} & \psnrcell{15.270} & \ssimcell{0.3460} & \lpipscell{0.4663} & \niqecell{4.585} & \psnrcell{19.300} & \ssimcell{0.6190} & \lpipscell{0.3765} & \niqecell{4.585} \\
6 & 70.7 & 0.2745 & \psnrcell{21.270} & \ssimcell{0.5899} & \lpipscell{0.3012} & \niqecell{3.586} & \psnrcell{15.325} & \ssimcell{0.3392} & \lpipscell{0.4830} & \niqecell{4.550} & \psnrcell{16.477} & \ssimcell{0.5628} & \lpipscell{0.5000} & \niqecell{5.024} \\
8 & 70.7 & 0.3529 & \psnrcell{19.884} & \ssimcell{0.5578} & \lpipscell{0.3249} & \niqecell{3.659} & \psnrcell{14.609} & \ssimcell{0.3464} & \lpipscell{0.4768} & \niqecell{4.123} & \psnrcell{15.225} & \ssimcell{0.5220} & \lpipscell{0.5557} & \niqecell{5.400} \\
\midrule

\multicolumn{15}{l}{\cellcolor[gray]{0.92}\textbf{Case 3 (Both vary):} $p$ and $\sigma$ vary together (same level index)} \\
0 & 1000.0 & 0.0392 & \psnrcell{26.938} & \ssimcell{0.7727} & \lpipscell{0.1263} & \niqecell{3.788} & \psnrcell{15.571} & \ssimcell{0.3724} & \lpipscell{0.4300} & \niqecell{4.567} & \psnrcell{21.642} & \ssimcell{0.7589} & \lpipscell{0.1826} & \niqecell{4.384} \\
2 & 308.0 & 0.1176 & \psnrcell{24.146} & \ssimcell{0.6841} & \lpipscell{0.2002} & \niqecell{3.859} & \psnrcell{15.105} & \ssimcell{0.3571} & \lpipscell{0.4614} & \niqecell{4.616} & \psnrcell{21.459} & \ssimcell{0.7031} & \lpipscell{0.2501} & \niqecell{4.834} \\
4 & 94.9 & 0.1961 & \psnrcell{22.563} & \ssimcell{0.6361} & \lpipscell{0.2707} & \niqecell{3.544} & \psnrcell{15.175} & \ssimcell{0.3382} & \lpipscell{0.4801} & \niqecell{4.791} & \psnrcell{20.131} & \ssimcell{0.6526} & \lpipscell{0.3415} & \niqecell{4.908} \\
6 & 29.2 & 0.2745 & \psnrcell{20.058} & \ssimcell{0.5713} & \lpipscell{0.3065} & \niqecell{3.679} & \psnrcell{15.116} & \ssimcell{0.3502} & \lpipscell{0.4634} & \niqecell{4.538} & \psnrcell{18.114} & \ssimcell{0.5725} & \lpipscell{0.4496} & \niqecell{5.463} \\
8 & 9.0 & 0.3529 & \psnrcell{19.111} & \ssimcell{0.5490} & \lpipscell{0.3383} & \niqecell{3.997} & \psnrcell{14.608} & \ssimcell{0.3345} & \lpipscell{0.4842} & \niqecell{4.378} & \psnrcell{13.919} & \ssimcell{0.4833} & \lpipscell{0.6104} & \niqecell{5.907} \\
\bottomrule
\end{tabular}%
}
\end{table}
\endgroup

\subsection{Spherical-Aberration Ablation}

For spherical aberration, we perform an ablation study by varying the aberration coefficient $c_{\mathrm{sph}}$ over multiple degradation levels. We synthesize five \PSF kernels representing progressively severe degradation: $c_{\mathrm{sph}} \in \{0.1\lambda, 0.25\lambda, 0.5\lambda, 1.0\lambda, 1.5\lambda\}$. As $c_{\mathrm{sph}}$ increases, energy diffuses from the central peak into peripheral diffraction rings, leading to significant blur and contrast loss. The performance of each model under these varying degrees of degradation is summarized in Table~\ref{tab:spherical_aberration}. Results suggest that agentic AI models are more reliable on mild degradations than on severe physics-driven blur, where inversion requires a more faithful understanding of the forward operator.
\begin{table}[htbp]
\caption{Deblurring performance under spherical aberration with two peak wavefront error levels $w_{\mathrm{err}}$.}
\label{tab:spherical_aberration}
\centering
\footnotesize
\setlength{\tabcolsep}{4pt}
\renewcommand{\arraystretch}{1.15}
\begin{tabular}{llcccc}
\toprule
Model & {$w_{\mathrm{err}}$ (waves)} & PSNR$\uparrow$ & SSIM$\uparrow$ & LPIPS$\downarrow$ & NIQE$\downarrow$ \\
\midrule

\multirow{2}{*}{Gemini}
& 0.10 & 23.233 & 0.6753 & 0.1731 & 3.8285 \\
& 0.20 & 22.153 & 0.6157 & 0.2364 & 3.8021 \\
\midrule

\multirow{2}{*}{GPT}
& 0.10 & 14.209 & 0.3825 & 0.4465 & 3.9218 \\
& 0.20 & 14.224 & 0.3883 & 0.4663 & 4.0804 \\
\midrule

\multirow{2}{*}{Qwen}
& 0.10 & 20.610 & 0.6652 & 0.2965 & 5.1592 \\
& 0.20 & -- & -- & -- & -- \\
\bottomrule
\end{tabular}
\end{table}

\subsection{API Cost}

We estimate the cloud-API cost of the benchmark by pairing the per-call token usage logged for every run with each provider's published prices; a full per-model breakdown is given in the supplementary material. The roughly 100-example-per-subtask scale keeps this diagnostic benchmark practical while still covering many computational-imaging operators. To compare the families on equal terms, all per-image figures use standard (non-batch) list pricing. At these rates the end-to-end cost per reconstructed image ranges from \$$0.048$ to \$$0.079$ depending on protocol and model (Table~\ref{tab:cost_main}): the Expert protocol runs a single editor call, whereas the Planner and Forward protocols add a planner call. The two hosted pipelines are close at standard rates---OpenAI is marginally lower per image---but Gemini becomes the cheaper pipeline once its $50\%$ batch discount is applied.

\begin{table}[t]
  \centering
  \caption{End-to-end API cost per reconstructed image by protocol, at standard list pricing (Gemini normalized from the $50\%$-off batch rates actually billed). Self-hosted Qwen is excluded as it incurs no per-call API charge.}
  \label{tab:cost_main}
  \small
  \begin{tabular}{lcc}
    \toprule
    Protocol & Gemini (\$/img) & OpenAI (\$/img) \\
    \midrule
    Expert (edit only)    & 0.069 & \textbf{0.048} \\
    Planner (plan + edit) & 0.077 & \textbf{0.062} \\
    Forward (plan + edit) & 0.079 & \textbf{0.075} \\
    \bottomrule
  \end{tabular}
\end{table}

The cost picture was not simply proprietary versus open. In our setup Gemini was the strongest overall image agent and also the best turnkey cost--performance option, while OpenAI was the least favorable among the hosted options. The self-hosted Qwen models incur no per-call API charge at all and remain attractive when local hosting or tighter deployment control matters, though they trail both hosted families in quality. The main tradeoff for Gemini was latency, typically a 10--20 minute batch turnaround rather than immediate responses, whereas Qwen was the most deployable, being locally hostable on commodity GPUs. Overall, Gemini was the best practical choice in our deployment setting, Qwen the most portable, and OpenAI the weakest on cost--performance.
 
\subsection{Failure Cases}

The main benchmark failure mode was reconstruction error on tasks that require explicit inversion of a sensing operator. Under the Expert protocol, the weakest families were lensless deconvolution, event reconstruction, and \CGH, with mean PSNR values of 7.96, 9.03, and 11.21 and LPIPS around 0.78--0.81. As discussed in Sec.~\ref{sec:results}, the models often produce plausible-looking images while missing the physically correct inverse solution.

At the same time, we also observed operational failures that were not captured by image metrics alone: some hosted models occasionally triggered safety filters or refused to return an answer, even when the prompt was part of a benign computational-imaging benchmark. These refusals were not the dominant scientific failure mode, but they still mattered in practice because they reduced reliability and added rerun overhead.

\begin{figure}[t]
	\centering
	\includegraphics[width=\linewidth]{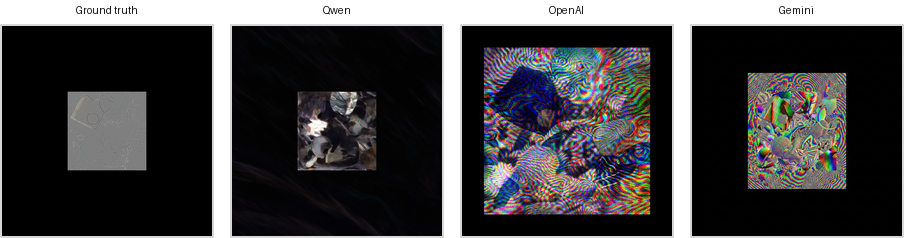}
	\caption{Representative \CGH failure case. The agents recover coarse structure, but none reconstructs the diffraction-consistent fine detail required by the task.}
	\label{fig:cgh_failure_cases}
\end{figure}

Figure~\ref{fig:cgh_failure_cases} shows a compact \CGH example preview. All three agents recover coarse structure, but none reconstructs the diffraction-consistent fine detail. The broader lesson is simple: current frontier imaging agents are competitive on familiar \ISP or restoration tasks, but they still break down when the task requires phase recovery, sparse inversion, or operator-aware reasoning, and proprietary systems can additionally fail at the API layer through filtering or refusal.

\subsection{Scope and Future Work}
This benchmark is intentionally a first version. We currently focus on image-centric agentic pipelines rather than more complex tool-using systems, so that the benchmark isolates imaging competence itself. We plan to open-source the evaluation infrastructure, host an online leaderboard, and expand the benchmark as frontier models evolve. Given the pace of progress in AI, we view \ImagingBench as a living benchmark rather than a fixed one-time release. A future version will focus on imaging-related coding tasks.

\section{Conclusion}

We introduced \ImagingBench, the first benchmark for evaluating agentic AI systems on computational imaging tasks spanning image reconstruction, sensing, optics, and calibration. Across 20 subtasks and three evaluation protocols, the results show a clear gap between high-level multimodal competence and physically grounded imaging understanding. Current agentic systems can often generate visually plausible outputs, but they remain consistently weaker than specialized non-agentic methods, especially on computational sensing tasks and other problems that require explicit inversion of a structured forward operator. Planner-guided prompting provides only modest gains over a strong fixed-prompt baseline, suggesting that the main limitation is not prompting alone but the underlying lack of low-level imaging competence. We hope \ImagingBench provides a useful foundation for measuring this gap and for guiding future work toward agentic models that not only see images, but also understand how images are formed and reconstructed.

\ifpeerreview \else
\section*{Acknowledgments}
This work used the Delta advanced computing and data resource at the National Center for Supercomputing Applications, supported by the U.S. National Science Foundation under award OAC 2005572 and by the State of Illinois. Delta is a joint effort of the University of Illinois Urbana-Champaign and its National Center for Supercomputing Applications.

This work also used Google Cloud Platform and Microsoft Azure through CloudBank under allocation CIS260120 from the Advanced Cyberinfrastructure Coordination Ecosystem: Services \& Support (ACCESS) program, supported by U.S. National Science Foundation grants \#2138259, \#2138286, \#2138307, \#2137603, and \#2138296. CloudBank is additionally supported by U.S. National Science Foundation grant \#1925001.
\fi

\bibliographystyle{IEEEtran}
\bibliography{references}

\begin{thebibliography}{10}
\providecommand{\url}[1]{#1}
\csname url@samestyle\endcsname
\providecommand{\newblock}{\relax}
\providecommand{\bibinfo}[2]{#2}
\providecommand{\BIBentrySTDinterwordspacing}{\spaceskip=0pt\relax}
\providecommand{\BIBentryALTinterwordstretchfactor}{4}
\providecommand{\BIBentryALTinterwordspacing}{\spaceskip=\fontdimen2\font plus
\BIBentryALTinterwordstretchfactor\fontdimen3\font minus
  \fontdimen4\font\relax}
\providecommand{\BIBforeignlanguage}[2]{{%
\expandafter\ifx\csname l@#1\endcsname\relax
\typeout{** WARNING: IEEEtran.bst: No hyphenation pattern has been}%
\typeout{** loaded for the language `#1'. Using the pattern for}%
\typeout{** the default language instead.}%
\else
\language=\csname l@#1\endcsname
\fi
#2}}
\providecommand{\BIBdecl}{\relax}
\BIBdecl

\bibitem{russakovsky2015imagenet}
O.~Russakovsky, J.~Deng, H.~Su, J.~Krause, S.~Satheesh, S.~Ma, Z.~Huang,
  A.~Karpathy, A.~Khosla, M.~Bernstein, A.~C. Berg, and L.~Fei-Fei, ``Imagenet
  large scale visual recognition challenge,'' \emph{International Journal of
  Computer Vision}, vol. 115, no.~3, pp. 211--252, 2015.

\bibitem{everingham2010pascal}
M.~Everingham, L.~Van~Gool, C.~K.~I. Williams, J.~Winn, and A.~Zisserman, ``The
  {PASCAL} visual object classes ({VOC}) challenge,'' \emph{International
  Journal of Computer Vision}, vol.~88, no.~2, pp. 303--338, 2010.

\bibitem{lin2014coco}
T.-Y. Lin, M.~Maire, S.~Belongie, J.~Hays, P.~Perona, D.~Ramanan,
  P.~Doll{\'a}r, and C.~L. Zitnick, ``Microsoft {COCO}: Common objects in
  context,'' in \emph{European Conference on Computer Vision (ECCV)}, 2014, pp.
  740--755.

\bibitem{wang2018glue}
A.~Wang, A.~Singh, J.~Michael, F.~Hill, O.~Levy, and S.~R. Bowman, ``{GLUE}: A
  multi-task benchmark and analysis platform for natural language
  understanding,'' in \emph{Proceedings of the 2018 {EMNLP} Workshop
  BlackboxNLP: Analyzing and Interpreting Neural Networks for NLP}, 2018, pp.
  353--355.

\bibitem{srivastava2023bigbench}
\BIBentryALTinterwordspacing
A.~Srivastava, A.~Rastogi, A.~Rao \emph{et~al.}, ``Beyond the imitation game:
  Quantifying and extrapolating the capabilities of language models,''
  \emph{Transactions on Machine Learning Research}, 2023. [Online]. Available:
  \url{https://openreview.net/forum?id=uyTL5Bvosj}
\BIBentrySTDinterwordspacing

\bibitem{liu2024mmbench}
Y.~Liu, H.~Duan, Y.~Zhang, B.~Li, S.~Zhang, W.~Zhao, Y.~Yuan, J.~Wang, C.~He,
  Z.~Liu, K.~Chen, and D.~Lin, ``{MMBench}: Is your multi-modal model an
  all-around player?'' in \emph{European Conference on Computer Vision (ECCV)},
  2024, pp. 216--233.

\bibitem{yue2024mmmu}
X.~Yue, Y.~Ni, K.~Zhang, T.~Zheng, R.~Liu, G.~Zhang, S.~Stevens, D.~Jiang,
  W.~Ren, Y.~Sun, C.~Wei, B.~Yu, R.~Yuan, R.~Sun, M.~Yin, B.~Zheng, Z.~Yang,
  Y.~Liu, W.~Huang, H.~Sun, Y.~Su, and W.~Chen, ``{MMMU}: A massive
  multi-discipline multimodal understanding and reasoning benchmark for expert
  {AGI},'' in \emph{Proceedings of the IEEE/CVF Conference on Computer Vision
  and Pattern Recognition (CVPR)}, 2024, pp. 9556--9567.

\bibitem{liu2024agentbench}
\BIBentryALTinterwordspacing
X.~Liu, H.~Yu, H.~Zhang, Y.~Xu, X.~Lei, H.~Lai, Y.~Gu, H.~Ding, K.~Men,
  K.~Yang, S.~Zhang, X.~Deng, A.~Zeng, Z.~Du, C.~Zhang, S.~Shen, T.~Zhang,
  Y.~Su, H.~Sun, M.~Huang, Y.~Dong, and J.~Tang, ``{AgentBench}: Evaluating
  {LLMs} as agents,'' 2023. [Online]. Available:
  \url{https://arxiv.org/abs/2308.03688}
\BIBentrySTDinterwordspacing

\bibitem{mialon2024gaia}
\BIBentryALTinterwordspacing
G.~Mialon, C.~Fourrier, T.~Wolf, Y.~LeCun, and T.~Scialom, ``{GAIA}: A
  benchmark for general {AI} assistants,'' in \emph{International Conference on
  Learning Representations (ICLR)}, 2024. [Online]. Available:
  \url{https://openreview.net/forum?id=fibxvahvs3}
\BIBentrySTDinterwordspacing

\bibitem{chow2025physbench}
\BIBentryALTinterwordspacing
W.~Chow, J.~Mao, B.~Li, D.~Seita, V.~Guizilini, and Y.~Wang, ``Physbench:
  Benchmarking and enhancing vision-language models for physical world
  understanding,'' in \emph{ICLR}, 2025. [Online]. Available:
  \url{https://arxiv.org/abs/2501.16411}
\BIBentrySTDinterwordspacing

\bibitem{yao2025mmmg}
\BIBentryALTinterwordspacing
J.~Yao, Y.~Hu, Y.~Yi, B.~Han, S.~Feng, G.~Yang, B.~Wen, R.~Krishna, L.~L. Wang,
  Y.~Tsvetkov, N.~A. Smith, and B.~Zhu, ``Mmmg: a comprehensive and reliable
  evaluation suite for multitask multimodal generation,'' 2025. [Online].
  Available: \url{https://arxiv.org/abs/2505.17613}
\BIBentrySTDinterwordspacing

\bibitem{suo2023computational}
J.~Suo, W.~Zhang, J.~Gong, X.~Yuan, D.~J. Brady, and Q.~Dai, ``Computational
  imaging and artificial intelligence: The next revolution of mobile vision,''
  \emph{Proceedings of the IEEE}, vol. 111, no.~12, pp. 1607--1639, 2023.

\bibitem{wang2022differentiable}
C.~Wang, N.~Chen, and W.~Heidrich, ``do: A differentiable engine for deep lens
  design of computational imaging systems,'' \emph{IEEE Transactions on
  Computational Imaging}, vol.~8, pp. 905--916, 2022.

\bibitem{yang2024curriculum}
X.~Yang, Q.~Fu, and W.~Heidrich, ``Curriculum learning for ab initio deep
  learned refractive optics,'' \emph{Nature communications}, vol.~15, no.~1, p.
  6572, 2024.

\bibitem{richardson1972}
W.~H. Richardson, ``Bayesian-based iterative method of image restoration,''
  \emph{Journal of the Optical Society of America}, vol.~62, no.~1, pp. 55--59,
  1972.

\bibitem{lucy1974}
L.~B. Lucy, ``An iterative technique for the rectification of observed
  distributions,'' \emph{The Astronomical Journal}, vol.~79, pp. 745--754,
  1974.

\bibitem{rudin1992}
L.~I. Rudin, S.~Osher, and E.~Fatemi, ``Nonlinear total variation based noise
  removal algorithms,'' \emph{Physica D: Nonlinear Phenomena}, vol.~60, no.
  1-4, pp. 259--268, 1992.

\bibitem{zhang2017beyond}
K.~Zhang, W.~Zuo, Y.~Chen, D.~Meng, and L.~Zhang, ``Beyond a gaussian denoiser:
  Residual learning of deep cnn for image denoising,'' \emph{IEEE transactions
  on image processing}, vol.~26, no.~7, pp. 3142--3155, 2017.

\bibitem{zhang2018learning}
K.~Zhang, W.~Zuo, and L.~Zhang, ``Learning a single convolutional
  super-resolution network for multiple degradations,'' in \emph{Proceedings of
  the IEEE conference on computer vision and pattern recognition}, 2018, pp.
  3262--3271.

\bibitem{antol2015vqa}
S.~Antol, A.~Agrawal, J.~Lu, M.~Mitchell, D.~Batra, C.~L. Zitnick, and
  D.~Parikh, ``{VQA}: Visual question answering,'' in \emph{Proceedings of the
  IEEE International Conference on Computer Vision (ICCV)}, 2015, pp.
  2425--2433.

\bibitem{agrawal2019nocaps}
H.~Agrawal, K.~Desai, Y.~Wang, X.~Chen, R.~Jain, M.~Johnson, D.~Batra,
  D.~Parikh, S.~Lee, and P.~Anderson, ``nocaps: Novel object captioning at
  scale,'' in \emph{Proceedings of the IEEE/CVF International Conference on
  Computer Vision (ICCV)}, 2019, pp. 8948--8957.

\bibitem{yang2023medmnistv2}
J.~Yang, R.~Shi, D.~Wei, Z.~Liu, L.~Zhao, B.~Ke, H.~Pfister, and B.~Ni,
  ``{MedMNIST} v2: A large-scale lightweight benchmark for 2d and 3d biomedical
  image classification,'' \emph{Scientific Data}, vol.~10, no.~1, p.~41, 2023.

\bibitem{irvin2019chexpert}
J.~Irvin, P.~Rajpurkar, M.~Ko, Y.~Yu, S.~Ciurea-Ilcus, C.~Chute, H.~Marklund,
  B.~Haghgoo, R.~Ball, K.~Shpanskaya, J.~Seekins, D.~A. Mong, S.~S. Halabi,
  J.~K. Sandberg, R.~Jones, D.~B. Larson, C.~P. Langlotz, B.~N. Patel, M.~P.
  Lungren, and A.~Y. Ng, ``{CheXpert}: A large chest radiograph dataset with
  uncertainty labels and expert comparison,'' in \emph{Proceedings of the AAAI
  Conference on Artificial Intelligence}, vol.~33, no.~1, 2019, pp. 590--597.

\bibitem{johnson2019mimiccxr}
A.~E.~W. Johnson, T.~J. Pollard, S.~J. Berkowitz, N.~R. Greenbaum, M.~P.
  Lungren, C.-y. Deng, R.~G. Mark, and S.~Horng, ``{MIMIC-CXR}, a de-identified
  publicly available database of chest radiographs with free-text reports,''
  \emph{Scientific Data}, vol.~6, p. 317, 2019.

\bibitem{nguyen2022vindrcxr}
H.~Q. Nguyen, K.~Lam, L.~T. Le, H.~H. Nguyen, H.~H. Pham, H.~Tong, D.~Dinh,
  D.~Nguyen, M.~Dao, V.~Vu \emph{et~al.}, ``{VinDr-CXR}: An open dataset of
  chest x-rays with radiologist's annotations,'' \emph{Scientific Data},
  vol.~9, no.~1, p. 429, 2022.

\bibitem{menze2015brats}
B.~H. Menze, A.~Jakab, S.~Bauer, J.~Kalpathy-Cramer, K.~Farahani, J.~Kirby,
  Y.~Burren, N.~Porz, J.~Slotboom, R.~Wiest \emph{et~al.}, ``The multimodal
  brain tumor image segmentation benchmark ({BRATS}),'' \emph{IEEE Transactions
  on Medical Imaging}, vol.~34, no.~10, pp. 1993--2024, 2015.

\bibitem{lau2018vqarad}
J.~J. Lau, S.~Gayen, A.~Ben~Abacha, and D.~Demner-Fushman, ``A dataset of
  clinically generated visual questions and answers about radiology images,''
  \emph{Scientific Data}, vol.~5, p. 180251, 2018.

\bibitem{liu2021slake}
B.~Liu, L.-M. Zhan, L.~Xu, L.~Ma, Y.~Yang, and X.-M. Wu, ``{SLAKE}: A
  semantically-labeled knowledge-enhanced dataset for medical visual question
  answering,'' in \emph{2021 IEEE 18th International Symposium on Biomedical
  Imaging (ISBI)}, 2021, pp. 1650--1654.

\bibitem{he2020pathvqa}
\BIBentryALTinterwordspacing
X.~He, Y.~Zhang, L.~Mou, E.~Xing, and P.~Xie, ``{PathVQA}: 30000+ questions for
  medical visual question answering,'' 2020. [Online]. Available:
  \url{https://arxiv.org/abs/2003.10286}
\BIBentrySTDinterwordspacing

\bibitem{zhang2023pmcvqa}
\BIBentryALTinterwordspacing
X.~Zhang, C.~Wu, Z.~Zhao, W.~Lin, Y.~Zhang, Y.~Wang, and W.~Xie, ``{PMC-VQA}:
  Visual instruction tuning for medical visual question answering,'' 2023.
  [Online]. Available: \url{https://arxiv.org/abs/2305.10415}
\BIBentrySTDinterwordspacing

\bibitem{chen2024gmai_mmbench}
\BIBentryALTinterwordspacing
P.~Chen, J.~Ye, G.~Wang, Y.~Li, Z.~Deng, W.~Li, T.~Li, H.~Duan, Z.~Huang,
  Y.~Su, B.~Wang, S.~Zhang, B.~Fu, J.~Cai, B.~Zhuang, E.~J. Seibel, Y.~Qiao,
  and J.~He, ``{GMAI-MMBench}: A comprehensive multimodal evaluation benchmark
  towards general medical ai,'' in \emph{Advances in Neural Information
  Processing Systems}, vol.~37, 2024. [Online]. Available:
  \url{https://proceedings.neurips.cc/paper_files/paper/2024/hash/ab7e02fd60e47e2a379d567f6b54f04e-Abstract-Datasets_and_Benchmarks_Track.html}
\BIBentrySTDinterwordspacing

\bibitem{luo2025mmmg}
\BIBentryALTinterwordspacing
Y.~Luo, Y.~Yuan, J.~Chen, H.~Cai, Z.~Yue, Y.~Yang, F.~Z. Daha, J.~Li, and
  Z.~Lian, ``{MMMG}: A massive, multidisciplinary, multi-tier generation
  benchmark for text-to-image reasoning,'' 2025. [Online]. Available:
  \url{https://arxiv.org/abs/2506.10963}
\BIBentrySTDinterwordspacing

\bibitem{chang2018hybrid}
J.~Chang, V.~Sitzmann, X.~Dun, W.~Heidrich, and G.~Wetzstein, ``Hybrid
  optical-electronic convolutional neural networks with optimized diffractive
  optics for image classification,'' \emph{Scientific reports}, vol.~8, no.~1,
  p. 12324, 2018.

\bibitem{yang2025vlmir}
\BIBentryALTinterwordspacing
C.~Yang, R.~Dong, and K.-M. Lam, ``Vision-language model guided image
  restoration,'' 2025. [Online]. Available:
  \url{https://arxiv.org/abs/2512.17292}
\BIBentrySTDinterwordspacing

\bibitem{geng2026optiagent}
\BIBentryALTinterwordspacing
Y.~Geng, L.~Sun, Y.~Gao, X.~Hu, Z.~Yi, X.~Qian, W.~Hu, J.~Bai, and K.~Wang,
  ``Optiagent: A physics-driven agentic framework for automated optical
  design,'' 2026. [Online]. Available: \url{https://arxiv.org/abs/2602.23761}
\BIBentrySTDinterwordspacing

\bibitem{shi2021towards}
L.~Shi, B.~Li, C.~Kim, P.~Kellnhofer, and W.~Matusik, ``Towards real-time
  photorealistic 3d holography with deep neural networks,'' \emph{Nature}, vol.
  591, no. 7849, pp. 234--239, 2021.

\bibitem{gao2025unified}
Q.~Gao, P.~Duan, H.~Lou, M.~Teng, Z.~Cai, X.~Chen, and B.~Shi, ``Unified
  reconstruction of static and dynamic scenes from events,'' in
  \emph{Proceedings of the Computer Vision and Pattern Recognition Conference},
  2025, pp. 27\,914--27\,923.

\bibitem{qiao2022depth}
X.~Qiao, C.~Ge, P.~Deng, H.~Wei, M.~Poggi, and S.~Mattoccia, ``Depth
  restoration in under-display time-of-flight imaging,'' \emph{IEEE
  transactions on pattern analysis and machine intelligence}, vol.~45, no.~5,
  pp. 5668--5683, 2022.

\bibitem{monakhova2019learned}
K.~Monakhova, J.~Yurtsever, G.~Kuo, N.~Antipa, K.~Yanny, and L.~Waller,
  ``Learned reconstructions for practical mask-based lensless imaging,''
  \emph{Optics express}, vol.~27, no.~20, pp. 28\,075--28\,090, 2019.

\bibitem{afifi2019color}
M.~Afifi, B.~Price, S.~Cohen, and M.~S. Brown, ``When color constancy goes
  wrong: Correcting improperly white-balanced images,'' in \emph{Proceedings of
  the IEEE/CVF conference on computer vision and pattern recognition}, 2019,
  pp. 1535--1544.

\bibitem{a2021beyond}
S.~A~Sharif, R.~A. Naqvi, and M.~Biswas, ``Beyond joint demosaicking and
  denoising: An image processing pipeline for a pixel-bin image sensor,'' in
  \emph{Proceedings of the IEEE/CVF conference on computer vision and pattern
  recognition}, 2021, pp. 233--242.

\bibitem{hdrplus_hasinoff2016burst}
S.~W. Hasinoff, D.~Sharlet, R.~Geiss, A.~Adams, J.~T. Barron, F.~Kainz,
  J.~Chen, and M.~Levoy, ``Burst photography for high dynamic range and
  low-light imaging on mobile cameras,'' \emph{ACM Transactions on Graphics
  (Proc. SIGGRAPH Asia)}, vol.~35, no.~6, 2016.

\bibitem{chen2015fpa}
H.~Chen, M.~Salman~Asif, A.~C. Sankaranarayanan, and A.~Veeraraghavan,
  ``Fpa-cs: Focal plane array-based compressive imaging in short-wave
  infrared,'' in \emph{Proceedings of the IEEE Conference on Computer Vision
  and Pattern Recognition}, 2015, pp. 2358--2366.

\bibitem{suvorov2022resolution}
R.~Suvorov, E.~Logacheva, A.~Mashikhin, A.~Remizova, A.~Ashukha, A.~Silvestrov,
  N.~Kong, H.~Goka, K.~Park, and V.~Lempitsky, ``Resolution-robust large mask
  inpainting with fourier convolutions,'' in \emph{Proceedings of the IEEE/CVF
  winter conference on applications of computer vision}, 2022, pp. 2149--2159.

\bibitem{google2026gemini31flash}
{Google DeepMind}, ``Gemini 3.1 flash-image model card,''
  \url{https://deepmind.google/models/model-cards/gemini-3-1-flash-image/},
  2026, accessed: 2026-02-26.

\bibitem{openai2026gptimage15}
\BIBentryALTinterwordspacing
{OpenAI}, \emph{GPT-Image 1.5 Model Documentation}, OpenAI, San Francisco, CA,
  2026, accessed: 2026-02-26. [Online]. Available:
  \url{https://developers.openai.com/api/docs/models/gpt-image-1.5}
\BIBentrySTDinterwordspacing

\bibitem{wu2025qwenimagetechnicalreport}
\BIBentryALTinterwordspacing
C.~Wu, J.~Li, J.~Zhou, J.~Lin, K.~Gao, K.~Yan, S.~ming Yin, S.~Bai, X.~Xu,
  Y.~Chen, Y.~Chen, Z.~Tang, Z.~Zhang, Z.~Wang, A.~Yang, B.~Yu, C.~Cheng,
  D.~Liu, D.~Li, H.~Zhang, H.~Meng, H.~Wei, J.~Ni, K.~Chen, K.~Cao, L.~Peng,
  L.~Qu, M.~Wu, P.~Wang, S.~Yu, T.~Wen, W.~Feng, X.~Xu, Y.~Wang, Y.~Zhang,
  Y.~Zhu, Y.~Wu, Y.~Cai, and Z.~Liu, ``Qwen-image technical report,'' 2025.
  [Online]. Available: \url{https://arxiv.org/abs/2508.02324}
\BIBentrySTDinterwordspacing

\bibitem{SIDD_2018_CVPR}
A.~Abdelhamed, S.~Lin, and M.~S. Brown, ``A high-quality denoising dataset for
  smartphone cameras,'' in \emph{IEEE Conference on Computer Vision and Pattern
  Recognition (CVPR)}, June 2018.

\bibitem{li2008image}
X.~Li, B.~Gunturk, and L.~Zhang, ``Image demosaicing: A systematic survey,'' in
  \emph{Visual Communications and Image Processing 2008}, vol. 6822.\hskip 1em
  plus 0.5em minus 0.4em\relax SPIE, 2008, pp. 489--503.

\bibitem{gopro_Nah_2017_CVPR}
S.~Nah, T.~H. Kim, and K.~M. Lee, ``Deep multi-scale convolutional neural
  network for dynamic scene deblurring,'' in \emph{CVPR}, July 2017.

\bibitem{stanford_lightfield}
{Stanford Computer Graphics Laboratory}, ``The stanford light field archive
  (2016),'' \url{https://lightfields.stanford.edu/LF2016.html}, 2016, accessed:
  2026-02-26.

\bibitem{div2k_Agustsson_2017_CVPR_Workshops}
E.~Agustsson and R.~Timofte, ``Ntire 2017 challenge on single image
  super-resolution: Dataset and study,'' in \emph{Proceedings of the IEEE
  Conference on Computer Vision and Pattern Recognition (CVPR) Workshops}, July
  2017.

\bibitem{fivek}
V.~Bychkovsky, S.~Paris, E.~Chan, and F.~Durand, ``Learning photographic global
  tonal adjustment with a database of input / output image pairs,'' in
  \emph{The Twenty-Fourth IEEE Conference on Computer Vision and Pattern
  Recognition}, 2011.

\bibitem{wu2019phasecam3d}
Y.~Wu, V.~Boominathan, H.~Chen, A.~Sankaranarayanan, and A.~Veeraraghavan,
  ``Phasecam3d—learning phase masks for passive single view depth
  estimation,'' in \emph{2019 IEEE International Conference on Computational
  Photography (ICCP)}.\hskip 1em plus 0.5em minus 0.4em\relax IEEE, 2019, pp.
  1--12.

\bibitem{wei2020physics}
K.~Wei, Y.~Fu, J.~Yang, and H.~Huang, ``A physics-based noise formation model
  for extreme low-light raw denoising,'' in \emph{Proceedings of the IEEE/CVF
  Conference on Computer Vision and Pattern Recognition}, 2020, pp. 2758--2767.

\bibitem{afifi2020deepWB}
M.~Afifi and M.~S. Brown, ``Deep white-balance editing,'' in \emph{Proceedings
  of the IEEE Conference on Computer Vision and Pattern Recognition}, 2020.

\bibitem{zhang2018ffdnet}
K.~Zhang, W.~Zuo, and L.~Zhang, ``Ffdnet: Toward a fast and flexible solution
  for cnn-based image denoising,'' \emph{IEEE Transactions on Image
  Processing}, vol.~27, no.~9, pp. 4608--4622, 2018.

\bibitem{Tsai2022BANet}
F.-J. Tsai, Y.-T. Peng, C.-C. Tsai, Y.-Y. Lin, and C.-W. Lin, ``Banet: A
  blur-aware attention network for dynamic scene deblurring,'' \emph{IEEE
  Transactions on Image Processing}, vol.~31, pp. 6789--6799, 2022.

\bibitem{wang2018esrgan}
X.~Wang, K.~Yu, S.~Wu, J.~Gu, Y.~Liu, C.~Dong, Y.~Qiao, and C.~Change~Loy,
  ``Esrgan: Enhanced super-resolution generative adversarial networks,'' in
  \emph{European Conference on Computer Vision (ECCV)}, 2018, pp. 0--0.

\end{thebibliography}

\end{document}